# SENDER: SEmi-Nonlinear Deep Efficient Reconstructor for Extraction Canonical, Meta, and Sub Functional Connectivity in the Human Brain

**Wei Zhang**
Augusta University
wzhang2@augusta.edu

**Yu Bao**
James Madison University
bao2yx@jmu.edu

## Abstract

Deep Linear and Nonlinear learning methods have already been vital machine learning methods for investigating the hierarchical features such as functional connectivity in the human brain via functional Magnetic Resonance signals; however, there are three major shortcomings: 1). For deep linear learning methods, although the identified hierarchy of functional connectivity is easily explainable, it is challenging to reveal more hierarchical functional connectivity; 2). For deep nonlinear learning methods, although non-fully connected architecture reduces the complexity of neural network structures that are easy to optimize and not vulnerable to overfitting, the functional connectivity hierarchy is difficult to explain; 3). Importantly, it is challenging for Deep Linear/Nonlinear methods to detect meta and sub-functional connectivity even in the shallow layers; 4). Like most conventional Deep Nonlinear Methods, such as Deep Neural Networks, the hyperparameters must be tuned manually, which is time-consuming.

Thus, in this work, we propose a novel deep hybrid learning method named SEmi-Nonlinear Deep Efficient Reconstruction (SENDER), to overcome the aforementioned shortcomings: 1). SENDER utilizes a multiple-layer stacked structure for the linear learning methods to detect the canonical functional connectivity; 2). SENDER implements a non-fully connected architecture conducted for the nonlinear learning methods to reveal the meta-functional connectivity through shallow and deeper layers; 3). SENDER incorporates the proposed background components to extract the sub-functional connectivity; 4). SENDER adopts a novel rank reduction operator to implement the hyperparameters tuning automatically.

To further validate the effectiveness, we compared SENDER with four peer methodologies using real functional Magnetic Resonance Imaging data for the human brain. Furthermore, the validation results show that SENDER outperforms the investigated methodologies regarding the reconstruction of identifiable canonical, meta, and sub-functional connectivity in the human brain, as well as efficiency and identifiability.

# 1      Introduction

The hierarchy of functional organization in the human brain [1-4] has been revealed by multiple deep linear machine learning techniques, such as Low-to High-Dimensional Independent Components Analysis (DICA) [5], Sparse Deep Dictionary Learning (SDDL) [6], Deep Non-negative Matrix Factorization (DNMF) [7], [8]. In addition, with the development of deep learning methods, a variety of deep nonlinear methods known as Deep Neural Networks (DNNs), e.g., Deep Convolutional Auto Encoder (DCAE), Restricted Boltzmann Machine (RBM), Deep Belief Network (DBN), and Convolutional Neural Network (CNN) [9-15], have been applied to discovering the hierarchical spatial features in brain, i.e., functional connectivity (FC), using functional Magnetic Resonance Imaging (fMRI). For instance, the RBM can extract hierarchical temporal features and effectively reconstruct FC networks with high accuracy [16, 17]. Furthermore, other latest research works have found the reasonable hierarchical temporal organization of task-based fMRI time series, each with corresponding task-evoked FCs [9, 10, 16, 17] using DCAE, RBM, and DBN. These deep learning techniques are generally considered deep nonlinear methods, e.g., DNNs, constructed with nonlinear activation functions, e.g., Sigmoid and/or Rectified Linear Unit (ReLU) [18].

From a neuroscience perspective, the reveal of FCs by methodologies can be divided into two folds: 1). Deep Linear methods, e.g., DNMF, and SDDL, usually focus on identifying meta-FCs explained as the recombination of canonical FCs reported by Smith in 2009 [19]; 2). On the other hand, Deep Nonlinear methods, e.g., DNNs, concentrate on investigating both meta- and sub-FCs, i.e., the lower-level features. Due to the utilization of the nonlinear activation functions, Deep Nonlinear Methods usually have a more substantial perception of meta-FCs than Deep Linear methods [12], [14]. However, these meta-FCs are often detected at the deeper layer in most Deep Linear/Nonlinear methods and are more difficult to interpret and/or explain [12], [14].

Therefore, we aim to propose a more advanced deep hybrid method that can benefit from both Deep Linear and nonlinear methods to simultaneously discover more identifiable canonical, meta, and sub-FCs from the shallow and deep layers. From a technical perspective, *SENDER* inherits the vital advantage of a deep linear method that it is easy to be optimized and does not require large training samples due to the convex target function, and the advantage of the deep nonlinear method such as more substantial perception using activation functions, and efficient non-fully connected architectures; in contrast, a fully connected structure in deep nonlinear methods enables the substantial perception but also involves in the difficult training issues, requiring large training samples to avoid overfitting and advanced optimizer to search the global optimum.

Furthermore, a more advanced deep hybrid method should overcome other shortcomings of deep nonlinear learning methods, such as DBN, Deep Boltzmann Machine (DBM), and DCAE, such as 1) large training samples [19-22]; 2) extensive computational resources, e.g., graphics processing units (GPUs) or tensor processing units (TPUs) [10, 11, 13]; 3) manual tuning of all hyperparameters, e.g., the number of layers and size of a dictionary, whereas some parameters such as sparse trade-off and step-length, are not denoted as hyperparameters [8-10]; 4) time-consuming training process [13, 14]; 5) uncertainty of convergence to the global optimum [13, 14, 19, 20, 22]; and 6) "black box" results that are challenging to explain [13, 14, 19].

*Contributions*. To be more specific, as follow:

1). *Accurate Approximation to Original Input*. We have proved that SENDER provides a comparable accuracy for approximation to original inputs to DNNs. Theorem 1.1 in Appendix A, Supplementary Material, presents the conclusion and proof details.

2). *Advanced Hybrid Architecture*. Unlike any deep linear or nonlinear method, SENDER employs a deep linear and nonlinear method with non-fully connected architectures, which provides an opportunity to explore the meta-FCs and the sub-FCs in the brain by introducing the background feature matrix. Theorem 1.2 in Appendix A, Supplementary Material, discusses the theoretical analysis of the advantages of the proposed deep hybrid architecture for learning.

3). *Fast Convergence Rate*. Given *STORM* [26] is used as an efficient optimizer to update all variables of *SENDER*, our theoretical analysis demonstrates that *SENDER* can maintain the same convergence rate as STORM itself, according to Theorem 1.3, Appendix A, Supplementary Material. Moreover, since the dimension of a dataset is continuously increasing, we also discuss the convergence of SENDER in an infinite dimensionality space, where all proofs can be found in Corollary 1.4 in Appendix A, Supplementary Material.

4). *Automatic Hyperparameters Tuning*. To implement the automatic tuning of hyperparameters, e.g., number of layers and number of units/neurons in each layer [23, 24], we develop a rank reduction technique named rank reduction operator (*RRO*) for *SENDER*. Specifically, *RRO* utilizes the orthogonal decomposition, e.g., *QR* decomposition, to estimate the rank of feature matrices, i.e., the number of units, via the weighted ratio (*WR*), the weighted difference (*WD*), and the weighted correlation (*WC*). These three techniques can consistently reduce the size of feature matrices through all layers until the estimated number of features equals one, suggesting the completion of decomposition. Thus, due to the generalization and efficacy of *QR* decomposition, *RRO* can tune all hyperparameters faster than Singular Value Decomposition (*SVD*) adopted by Principal Component Analysis (PCA) [23, 24]. The details of *RRO* implementation can be viewed as Algorithms 2.3 to 2.5, Appendix B, in Supplementary Material. In addition, the theoretical analytics of *RRO* can be viewed as Theorem 2.1, included in Appendix B, Supplementary Material.

4). *Reduced Accumulative Error via Matrix Backpropagation (MBP)*. Given that the accumulative error could potentially deteriorate the reconstruction accuracy, we utilize a technique *MBP* [6, 7, 25] to reduce the accumulative error. The implementation of *MBP* can be found in Algorithm 2.5 in Appendix B, Supplementary Material.

**Related Works and Methodological Validation**. *SENDER* is validated on real resting-state fMRI signals and compared with four other peer methods. The validation results show that *SENDER* can detect more identifiable canonical, meta, and sub-FCs in the human brain than the representative deep linear/nonlinear methods and is easier to be optimized than DNNs based on our theoretical analyses in Theorem 1.2. Furthermore, unlike most deep learning methods, some meta-FCs can be derived from *SENDER* even at shallow layers. Moreover, sub-FCs can be directly extracted from the background component matrix, whereas minor features, e.g., sub-FCs, are detected through shallow to deeper layers in most deep nonlinear methods.

## 2 Method

This section provides the details of *SENDER*, including optimizer, optimization function, automatic hyperparameters tuning technique, MBP, theoretical analysis of the convergence rate, and approximate accuracy.

### 2.1 Semi-Estimated Nonlinear Deep Efficient Reconstructor

As introduced, SENDER employs a hybrid architecture of Deep Linear/Nonlinear methods with an efficient non-fully connected architecture and the nonlinear activation function used in DNNs. In detail, the optimization function governing *SENDER* is shown below:

$$min_{S_i \in \mathbb{R}^{m \times n}} \bigcup_{i=1}^{k} \|S_i\|_1$$

$$s.t. \forall k \in [2, M], \left(\prod_{i=1}^{k} X_i\right) \cdot Y_k + \left(\prod_{i=1}^{k} U_i\right) \cdot \mathcal{N}_k \cdot V_k + S_k = I$$

$$\left(\prod_{i=1}^{j} X_i\right) \cdot Y_j + \left(\prod_{i=1}^{j} U_i\right) \cdot \mathcal{N}_j \cdot V_j + S_j \leftarrow I - S_{j-1}, \forall\ 1 \leq i < j \leq k$$

(1)

where $\{X_i\}_{i=1}^{k}$ represents the hierarchical weight matrices or mixing matrices of the linear method, for instance, $X_i$ indicates the weight matrix of linear method of at the $i^{th}$ layer, and $k$ denotes the total number of layers. similarly, $\{U_i\}_{i=1}^{k}$ denotes the weight matrices for nonlinear method at $i^{th}$ layer. Furthermore, $\{Y_i\}_{i=1}^{k}$ represents the *canonical or meta-FCs* derived via linear method; for instance, $Y_i$ indicates the canonical or meta-FCs of the $i^{th}$ layer; and meanwhile $\{V_i\}_{i=1}^{k}$ defines the meta-FCs revealed via the nonlinear method. Furthermore, $\{S_i\}_{i=1}^{k}$ is a set of matrices that represent the background components, which are denoted as sub-FCs, due to their sparsity. Moreover, $\mathcal{N}_i$ represents the nonlinear activation function at the $i^{th}$ layer, e.g., Sigmoid or Rectified Linear Unit (ReLU) [14, 20]. And, if we assume the total number of layers as $M$, the original input data $I$ can be decomposed following: $(\prod_{i=1}^{M} X_i) \cdot Y_M + (\prod_{i=1}^{M} U_i) \cdot \mathcal{N}_M \cdot V_M + S_M$.

As shown in Eq. (1), our fundamental assumption is the previously revealed FCs, e.g., $Y_{i-1}$ or $V_{i-1}$, can be decomposed further as a linear product of a deeper weight matrix $X_i$ and FCs as $Y_i$ or a nonlinear representation of a deeper weight matrix $U_i$ and FCs as $V_i$, respectively. In addition, the optimization function governing *SENDER* consists of more variables than conventional deep linear/nonlinear methods, such as *DICA*, *DNMF*, and *SDDL*.

Before optimizing Eq. (1), we can convert it into an augmented Lagrangian function. If considering the $k^{th}$ layer, we have:

$$\mathcal{L}_\rho(\prod_{i=1}^{k} X_i, Y_k, S_k, \mathcal{N}_k) \stackrel{\text{def}}{=} \frac{\rho}{2} \left\| I - \left(\prod_{i=1}^{k} X_i\right) \cdot Y_k - \left(\prod_{i=1}^{k} U_i\right) \cdot (\mathcal{N}_k \cdot V_k) \right\|_F^2 + \frac{1}{\rho} \|S_k\|_1 \quad (2)$$

The sparse trade-off controlling the sparsity of background components denoted as $\bigcup_{i=1}^{k} \|S_i\|_1$ is determined by $\frac{1}{\rho}$ that can also be estimated using Rose Algorithm [27]. Naturally, it is easier to employ alternative optimization strategies [25, 26] to optimize Eq. (2). Due to the efficacy of the recent reported *STORM* optimizer [27], we adopt *STORM* to update all the variables included in *SENDER*. And the $\ell_1$ norm of $S_k$ shown in Eq. (1) can be solved directly using the shrinkage method [28].

Denote *STORM* [26] as an operator $\mathcal{T}$. The iterative format of *STORM* using an alternative strategy of optimization to update all the variables in Eq. (2) can be presented as follows:

$$X_k^{it+1} \leftarrow \mathcal{T} \cdot X_k^{it} \quad (3\text{-}1)$$

$$Y_k^{it+1} \leftarrow \mathcal{T} \cdot Y_k^{it} \quad (3\text{-}2)$$

$$U_k^{it+1} \leftarrow \mathcal{T} \cdot U_k^{it} \quad (3\text{-}3)$$

$$V_k^{it+1} \leftarrow \mathcal{T} \cdot (\mathcal{N}_k \cdot V_k^{it}) \quad (3\text{-}4)$$

$$S_k^{it+1} \leftarrow Shrinkage\left[ I - \left(\prod_{i=1}^{k} X_k^{it}\right) \cdot Y_k^{it} - \left(\prod_{i=1}^{k} U_k^{it}\right) \cdot (\mathcal{N}_k \cdot V_k^{it}) \right] \quad (3\text{-}5)$$

In detail, in Eqs. (3-1) to (3-4), the current iteration is represented as *it*; for example, in Eq. (3-1), $X_k^{it}$ is updated by the optimizer STORM while $Y_k^{it}$, $U_k^{it}$, and $V_k^{it}$ are treated as constants; similar mechanism applies to $Y_k^{it}, U_k^{it}$, and $V_k^{it}$. Finally, Eq. (3-5) demonstrates the shrinkage and minimization of the background components, i.e., sub-FCs denoted as $S_k^{it+1}$.

## 2.2 Rank Reduction Operator for Automatic Tuning Hyperparameters

To implement the automatic hyperparameters tuning and reduce the high dimensionality of the original dataset, we introduce a novel technique named *RRO* which aims to calculate the rank of the current feature matrices, e.g., $\{Y_i\}_{i=1}^{k}$ and $\{V_i\}_{i=1}^{k}$ in *SENDER*, until the rank of the feature matrices are equivalent to one. More specifically, *RRO* performs rank-revealing by consistently using orthogonal decomposition via *QR* factorization to efficiently estimate the rank of feature matrices [23, 24]. Due to the effectiveness of *QR* factorization, *RRO* can be used to decompose sparse and overcomplete matrices. The mathematical formula of *RRO* is:

$$\mathcal{R} \begin{bmatrix} a_1 \\ a_2 \\ \vdots \\ a_{n-1} \\ a_n \end{bmatrix} = \begin{bmatrix} a_1^{(1)} \\ a_2^{(1)} \\ \vdots \\ a_{n-2}^{(1)} \\ a_{n-1}^{(1)} \end{bmatrix} \mathcal{R}^k \begin{bmatrix} a_1 \\ a_2 \\ \vdots \\ a_{n-1} \\ a_n \end{bmatrix} = \begin{bmatrix} a_1^{(1)} \\ a_2^{(1)} \\ \vdots \\ a_{n-k-1}^{(1)} \\ a_{n-k}^{(1)} \end{bmatrix} = [\hat{a}] \quad (4)$$

where operator $\mathcal{R}$ represents *RRO*, and $\{a_i\}_{i=1}^{n}$ is a series of vectors with $a_i$ representing a single vector. Assume *RRO* is repeatedly applied on a series of vectors $\{a_i\}_{i=1}^{n}$ for $k$ times, we have $rank(\mathcal{R}^k \cdot [a_1, a_2, \cdots, a_n]) < rank([a_1, a_2, \cdots, a_n])$ hold; furthermore, if $k$ is large enough, e.g., $\forall k > 0, \exists N \in \mathbb{N}$, when $k > N$, $rank(\mathcal{R}^k \cdot [a_1, a_2, \cdots, a_n]) = 1$ holds, thus $k$ is treated as the total number of layers since the rank of the feature matrix at $k^{th}$ layer equals one after utilizing *RRO* for $k$ times repeatedly. Moreover, we prove, if $\mathcal{R}: \mathbb{R}^{M \times N} \to \mathbb{R}^{M \times N}$, $M < \infty$, $N < \infty$, then we have $\|\mathcal{R}\| < \infty$, meaning the operator $\mathcal{R}$, i.e., *RRO* technique, is a bounded operator denoted in a finite-dimensional space. The detail of the proof can be viewed in Theorem 2.2, Appendix B, Supplementary Material.

Assume $r^*$ is the initially estimated rank and $r$ is the optimal rank estimation of the input signal matrix $I$, we have $r^* \geq r$; the diagonal of the upper-triangular matrix can be achieved after applying *QR* factorization on signal matrix $I$. In detail, at first, the diagonal of matrix $R$, derived from the feature matrix using *QR* decomposition, is non-increasing in magnitude [23, 24]; furthermore, along the main diagonal of matrix $R$, three techniques named *WR*, *WD*, and *WC* are applied to calculate the maximum rank shown in Eqs (5)-(7); then, $I$ is replaced by $I - S_k$, $k = 1, 2, \cdots, M$, iteratively. It indicates that the rank-reducing technique can yield a reasonable solution using *QR* factorization [23, 24]. The following formulas provide details of *WR*, *WD*, and *WC*.

Assume $d \in \mathbb{R}^{1 \times r}$ and $r \in \mathbb{R}^{r-1}$, then *WR* can be calculated by Eq. (5):

$$d_i \leftarrow |R_{ii}|$$
$$wr_i \leftarrow \frac{d_i}{d_{i+1}} \tag{5}$$

where $R_{ii}$ denotes a diagonal element of matrix $R$ calculated by *QR* decomposition and $wr_i$ is an element of *WR*. The value of each *WR* is derived by the ratio of the current element of diagonal and the next element as shown in Eq. (5).

Similarly, *WD* can be derived by:

$$wd_i \leftarrow \frac{|d_i - d_{i-1}|}{\sum_{k=1}^{i-1} d_k} \tag{6}$$

In Eq. (6), *WD* is defined as the absolute difference between the current diagonal element and the previous one divided by the cumulative sum of all the previous diagonal elements.

Furthermore, Eq. (7) describes the proposed *WC* as:

$$r_i \leftarrow |R_i|, 1 \leq i \leq n$$
$$wc_i \leftarrow \frac{|corr(r_{i-2}, r_{i-1}) - corr(r_{i-1}, r_i)|}{\sum_{k=i-2}^{i} \|r_k\|_2^2}, 3 \leq i \leq n \tag{7}$$

where $wc_i$ represents an element of *WC* and is the ratio of the absolute difference of three adjacent columns and the summed of all vectors' $\ell_2$ norm.

Thus, the *RRO* iteratively calculates the maximum value position from *WR*, *WD*, and *WC* to estimate the rank of R. The details of pseudocodes to implement *WR*, *WD*, and *WC* can be viewed in Algorithm 2.2, 2.3, and 2.4 in Appendix B, Supplementary Material.

## 2.3 Matrix Backpropagation

Another important technical contribution introduced in this work is *MBP* implemented to *SENDER* to further reduce the potential accumulative error after finishing the updates of all variables in *SENDER*. Assume the number of total layers as $M$, Eqs. (8)-(10) describe the details of MBP applied to the linear method part, and Eqs. (11)-(13) provide the mathematic formula of MBP for nonlinear method part in SENDER [31], [32], [42]. All variables, such as $X_k$, $Y_k$, $U_k$, $V_k$, and $I$ are denoted as the same in Section 2.1.

$$\hat{Y}_k \leftarrow Y_k, k = M \tag{8}$$
$$\hat{Y}_k \leftarrow X_{k+1} \hat{Y}_{k+1}, k < M$$

In addition, we denote the product of hierarchical weight matrices as $\psi$ in Eq. (9)

$$\psi \leftarrow \prod_{i=1}^{k} X_i \quad 1 < k < M \tag{9}$$

Then, the following equation describes the crucial steps of *MBP* to update variables of $\{X_i\}_{i=1}^{M}$, and $\{Y_i\}_{i=1}^{M}$ representing the hierarchical weight and feature matrices including all canonical and some meta-FCs, respectively.

$$Z_k \leftarrow I - \prod_{i=1}^{k} U_i \cdot (\mathcal{N}_k \cdot V_k) \tag{10-1}$$

$$\hat{Y}_k^+ \leftarrow Y_k^+ \odot \sqrt{\frac{[\psi^T Z_k]^+ + [\psi^T \psi]^- \hat{Y}_k^+}{[\psi^T Z_k]^- + [\psi^T \psi]^+ \hat{Y}_k^+}} \tag{10-2}$$

More details can be viewed in Algorithm 2.5, Appendix B, Supplementary Material.

Similarly, *MBP* is employed to further reduce the potential accumulative errors caused by the deep nonlinear method, [6, 7, 25]. Denote $U_k = \lim_{it \to \infty} U_k^{it}$, $X_k = \lim_{it \to \infty} X_k^{it}$, and $Y_k = \lim_{it \to \infty} Y_k^{it}$, we have:

$$K \leftarrow \left(\prod_{k=1}^{M} U_k\right)^T \cdot \left(I - \prod_{k=1}^{M} X_k \cdot Y_M\right) \tag{11-1}$$

$$P_k^{it} \leftarrow (U_k^{it})^T U_k^{it} \cdot \prod_{k=1}^{M-1} \max(U_k) \tag{11-2}$$

In Eqs. (11-1) and (11-2), there are two important variables, i.e., deep weight matrices and feature matrices in the nonlinear method, updated by the following backpropagation techniques shown from Eqs. (12-1) to (12-4) [6, 25]. The following equations employ $c_k^{it}$, $C_k^{it}$, $d_k^{it}$, and $D_k^{it}$ to perform MBP [6, 7] using the derivative of the inverse activation function $\frac{d\mathcal{N}^{-1}(s)}{ds}$:

$$c_k^{it} \leftarrow \max(U_{k-1}) \cdot \frac{d\mathcal{N}^{-1}(s)}{ds}, s = U_k^{it} V_k^{it} \tag{12-1}$$

$$d_k^{it} \leftarrow \frac{d\mathcal{N}^{-1}(s)}{ds}, s = U_k^{it} V_k^{it} \tag{12-2}$$

$$C_k^{it} \leftarrow (U_k^{it})^T \cdot (P_k^{it} \cdot \mathcal{N}^{-1}(s) - K) \odot c_k^{it}, s = U_k^{it} V_k^{it} \tag{12-3}$$

$$D_k^{it} \leftarrow (U_{k-1}^{it})^T \cdot (U_{k-1}^{it} \cdot \mathcal{N}^{-1}(s) - V_{k-1}^{it}) \odot d_k^{it} \cdot (V_k^{it})^T, s = U_k^{it} V_k^{it} \tag{12-4}$$

Finally, the following equations show the process of updating the weight matrix $U_k$ and the feature matrix $V_k$ in the $k^{th}$ layer. In addition, $T$ is a constant value determined as 0.01 [6, 7, 25].

$$V_k^{it+1} \leftarrow V_k^{it} - \frac{T}{2^{it}}(C_k^{it}) \tag{13-1}$$

$$U_k^{it+1} \leftarrow U_k^{it} - \frac{T}{2^{it}}(D_k^{it}) \tag{13-2}$$

## 2.4    Approximation and Convergence Rate of SENDER

In this section, we theoretically analyze the efficacy and discusses the performance of approximation and convergence rate of *SENDER*. Due to *SENDER* being organized as a composition of linear and nonlinear functions, the following theorem demonstrates that SENDER can approximate any real function that is almost-everywhere infinite [29] with high accuracy. The proof of Theorem 1.1 can be viewed in Appendix A, Supplementary Material.

***Theorem 1.1*** **(Accurate Approximation of SENDER)** Given a real function $f: \mathbb{R}^{P \times Q} \to \mathbb{R}^{P \times Q} \cup \{\pm \infty\}$ and $m(\{I \in \mathbb{R}^{P \times Q} : f(I) = \pm \infty\}) = 0$ where $m(\cdot)$ represents the Lebesgue measure [29]. Given $I \in \mathbb{R}^{P \times Q}$, *SENDER* includes a linear method and a nonlinear method with multiple activation functions denoted as $\{P_k(I)\}_{k=1}^{N_1}$ and $\{\mathcal{N}_k(I)\}_{k=1}^{N_2}$. If $P_k$ denotes a series of matrix polynomials and $\mathcal{N}_k$ denotes a smooth activation function, then we have $\forall\,\varepsilon > 0$, $N > 0$, $N_1 > N$, $N_2 < \infty$, $\left\|\{P_k(I)\}_{k=1}^{N_1} + \{\mathcal{N}_k(I)\}_{k=1}^{N_2} - f(I)\right\| \le \varepsilon$.

Theorem 1.1 demonstrates that SENDER enables an accurate approximation to the original input $f(I)$, even if it is almost-everywhere finite, e.g., $f(I) = \pm\infty, m(I) = 0$.

Since the fully-connected architecture is widely used in DNNs with various activation functions, the optimization function of the conventional neural network can be very complicated. Therefore, SENDER implements a non-fully connected architecture to reduce the complexity of network structures and thus improve the efficiency of optimization since the global optimum of the optimization function can be found by a gradient optimizer. The details of this conclusion are proved as Theorem 1.2 in Appendix A, supplementary material.

***Theorem 1.2*** **(Efficiency of Non-Fully Connected Architecture of SENDER)** Given a series of non-smoothed activation function $\{f_i\}_{i=1}^N$, defined on $[a,b] \subseteq \mathbb{R}^1$, assume $\{f_i\}_{i=1\,i}^N \subseteq Lip1([a,b]\backslash \{U(x_i,\delta)\})\ i \in \mathbb{N}$, and $U(x_i,\delta)$ is an open cubic with the center $x_i$ and radius $\delta > 0$. The composition of $f_i$, $f_j \in \{f_i\}_{i=1}^N$ are denoted as $f_{j,i} \stackrel{\text{def}}{=} f_j(f_i(x))$; the various composition $\mathcal{F} \stackrel{\text{def}}{=} f_{\cdots,k,\cdots j.i} \in Lip1([a,b]\backslash[c,d])$ holds, when $k \to \infty$, $[c,d] \supseteq \cup_{i=1}^k U(x_i,\delta)$ and $m([c,d]) \ne 0$; moreover, given $t \to \infty$, the summation of $\sum_{i=1}^t \mathcal{F}_i$ leads to $\sum_{i=1}^t \mathcal{F}_i \notin Lip1([a,b]\backslash[c',d'])$, $[c',d'] \supseteq [c,d]$, and $m([c',d']) \ne 0$. And $m(\cdot)$ represents the Lebesgue measure.

Theorem 1.2 demonstrates the infinite composition of activation function, such as fully connected and very DNN architecture, even with a single non-smooth point, which finally results in a non-smooth interval as $[c',d']$. Meanwhile, this theorem demonstrates that the non-fully connected architectures can be more easily optimized, i.e., the global optimum of the non-fully connected method is easier to be searched by a gradient-based optimizer.

The following definition and theorems support that SENDER can maintain the convergence rate of the original optimizer STORM. Moreover, we analyze the convergence rate of SENDER in the finite and infinite dimensional space, respectively.

***Definition 1.1*** **(SENDER Operator)** Denote the Random Initialization Operator as $\mathcal{P}: \mathbb{R}^{P \times Q} \to \mathbb{R}^{P \times Q}$, the Sparse Operator as $\mathcal{S}: \mathbb{R}^{P \times Q} \to \mathbb{R}^{P \times Q}$ and the STORM operator as $\mathcal{T}: \mathbb{R}^{P \times Q} \to \mathbb{R}^{P \times Q}$. Their norms can be represented as $\frac{1}{r} \stackrel{\text{def}}{=} \|\mathcal{P}\|$, $\frac{1}{s} \stackrel{\text{def}}{=} \|\mathcal{S}\|$, and $\frac{1}{t} \stackrel{\text{def}}{=} \|\mathcal{T}\|$; in Theorems 2.1 to 2.4 in Appendix B, Supplementary Material, we have: $0 < \frac{1}{s} < 1$ and $0 < \frac{1}{r}, \frac{1}{t} < \infty$, but $0 < \frac{1}{t^k} < 1$, $k < \infty$.

***Theorem 1.3*** **(Convergence Rate of SENDER in Finite Dimensionality Space)** Denote STORM as an operator $\mathcal{T}$ in a finite dimensionality space. Due to the convergence of STORM with a rate of $\mathcal{O}(T^{\frac{1}{2}} + \sigma^{\frac{1}{3}}/T^{\frac{1}{3}})$, the convergence of *SENDER* is the same as STORM.

Theorem 1.3 shows that SENDER can converge as fast as STORM due to the prerequisite of finite dimensionality space. We further prove the convergence of SENDER in an infinite dimensionality space in Collaroy 3.1.

***Collaroy 1.3*** **(Convergence of SENDER in Infinite Dimensionality Space)** Given the infinite dimensionality space [30, 31], denote SENDER as an operator $\mathcal{D}: \mathbb{R}^{\infty \times \infty} \to \mathbb{R}^{\infty \times \infty}$. Denote $\mathcal{D}$ as an infinite dimensional matrix operator with each element represented as $d_{i,j}$ and $r_{i,j} \in \mathbb{R}^{\infty \times \infty}$, $i,j \to \infty$, respectively. $\mathcal{D}$ can converge to a fixed point, if and only if $d_{i,j}^k \cdot r_{i,j}$ is $\mathcal{O}\left(\frac{1}{n^p}\right), n \in \mathbb{N}, p > 1, p \in \mathbb{R}$.

***Theorem 1.4*** **(Convergence of SENDER using Alternative Update)** Given $\{\mathcal{F}_{i,j,k,t}\}_{i,j,k,t=1}^{\infty}$, $\{\mathcal{C}_{j,k,t}\}_{j,k,t=1}^{\infty}$, $\{\mathcal{H}_{k,t}\}_{k,t=1}^{\infty}$, and $\{\mathcal{K}_t\}_{t=1}^{\infty}$, are series of continuous operators [30] applied on a finite dimensionality space, we have: $\mathcal{F}_{i,j,k,t}, \mathcal{C}_{j,k,t}, \mathcal{H}_{k,t}, \mathcal{K}_t: \mathbb{R}^{P \times Q} \to \mathbb{R}^{P \times Q}$. If $\lim_{i \to \infty} \mathcal{F}_{i,j} \to \mathcal{C}_{j,k,t}$, $\lim_{j \to \infty} \mathcal{C}_{j,k,t} \to \mathcal{H}_{k,t}$, $\lim_{k \to \infty} \mathcal{H}_{k,t} \to \mathcal{K}_t$, and $\lim_{t \to \infty} \mathcal{K}_t \to \mathcal{G}$, then, $\lim_{n \to \infty} \mathcal{F}_{i_n,j_n,k_n,t_n} \to \mathcal{G}$ holds.

Theorem 4.1 demonstrates that a computational model with multiple variables satisfying Corollary 1.2 can converge to a fixed point via an alternative strategy. The proof of Theorem 4.1 can be viewed in Appendix A, Supplementary Material. Moreover, in Theorem 4.1, due to the convexity of the Augmented Lagrange function [32], each independent approximation, such as $\lim_{n \to \infty} \mathcal{F}_{i_n, j_n, k_n, t_n} \to \mathcal{G}$ can converge to the operator.

## 3 Results

### 3.1 Comparison of Identified Canonical, Meta, and Sub-FCs via SENDER and Other Peer Four Methods

To validate SENDER, we employ the resting-state fMRI signals from all healthy individuals in Consortium for Neuropsychiatric Phenomics (CNP) (https://openfmri.org/dataset/ds000030/). To reduce the heterogeneous influence caused by parameter tuning, all hyperparameters are tuned the same as *SENDER*'s hyperparameter estimations. The estimated number of layers for SENDER is two. The sizes of the first and second layers are 40 and 10, respectively. In addition, other peer algorithms' parameters are tuned following in [5-8]. The activation function of all layers of SENDER and DBN is set as ReLU. Furthermore, all abbreviations of templates can be found in Table S1, Appendix A, Supplementary Material.

Figure 1 and 2 present the reconstruction of the canonical FCs via SENDER and other peer methods, e.g., DICA, DNMF, SDDL, and DBN; the identification results via the linear method in SENDER and other four peer algorithms are compared with canonical templates [19]. In short, this experimental validation demonstrates that the reconstruction of canonical FCs via SENDER is not significantly different from the canonical templates.

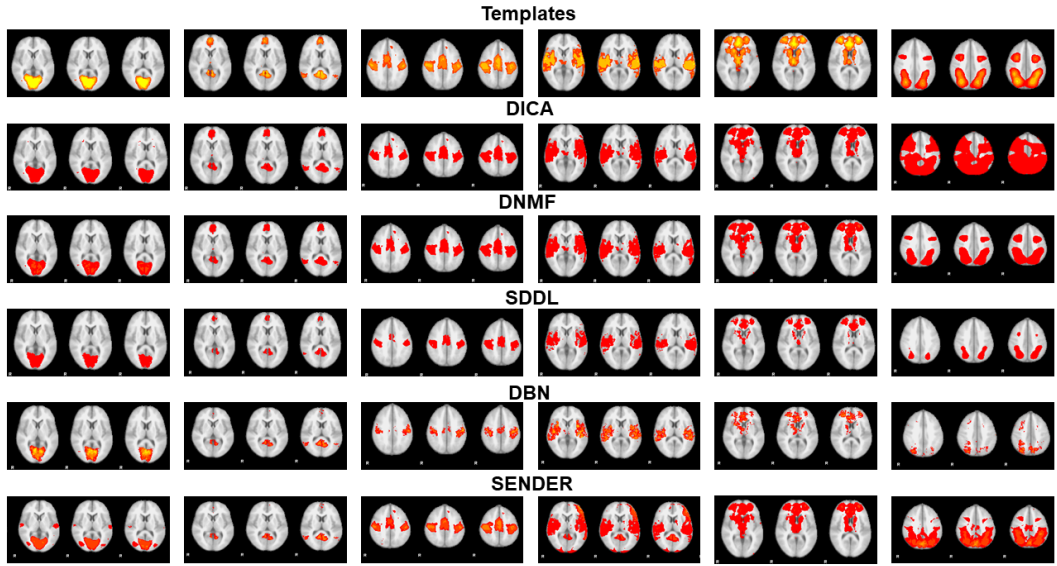

Figure 1. This figure presents three representative slices of reconstructed six canonical FCs via SENDER and six other peer methods.

Furthermore, the quantitative comparison of canonical FCs is provided in Figure 5 (b), where it compares the similarity of reconstructed canonical FCs derived via the linear method in SENDER and other four peer methods and publicly released canonical templates [19]. The similarity is measured by Hausdorff Distance [34]. In general, the similarity of identified canonical FCs via SENDER and canonical templates is higher/comparable to other peer algorithms.

Moreover, Figure 2 shows the reconstruction of meta-FCs derived by the nonlinear method in SENDER and the other four peer methods. Overall, the nonlinear methods in SENDER can successfully reconstruct meta-FCs with a higher similarity calculated with the meta-templates [41, 42].

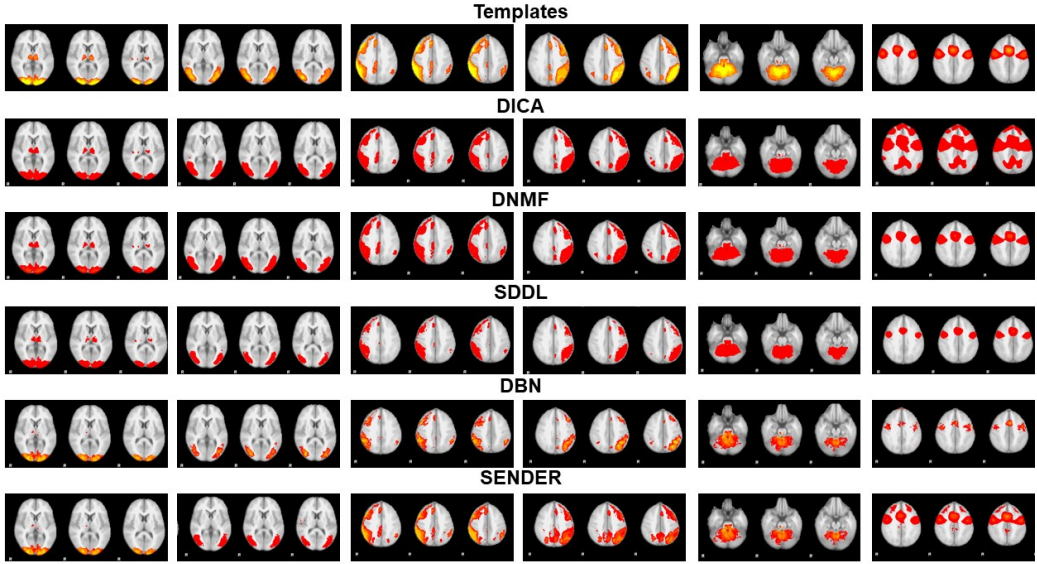

Figure 2. This figure presents the representative slices of reconstructed another six canonical FCs via SENDER and the other six peer methods.

In Figure 3, according to qualitative observation, the nonlinear method in SENDER and DBN can reconstruct meta-FCs more similarly to the templates; meanwhile, DICA and DNMF only reconstruct five and three FCs with higher similarity to the original templates, respectively. Furthermore, for SDDL, the reconstruction of FCs in the first and second columns is similar to the original templates.

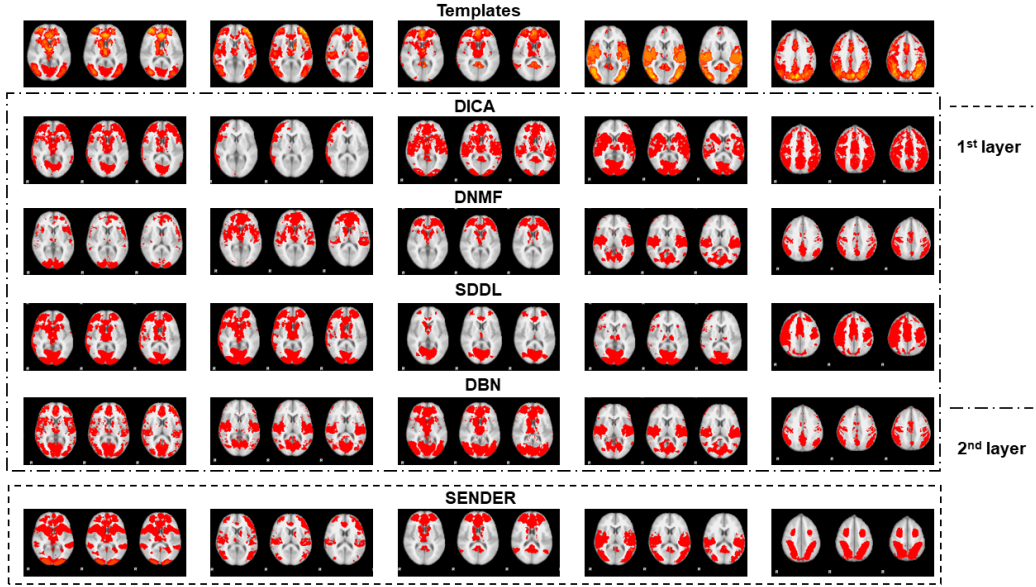

Figure 3. All qualitative comparisons revealed five meta-FCs at the first or second layer via the nonlinear method in *SENDER* and the other four peer methods. Please notice that the meta-FCs can be identified at the first and second layers using *SENDER*; nevertheless, the other four peer methods need to extract meta-FCs in the deeper layers.

In detail, for instance, *SENDER* can reveal meta-FCs using the nonlinear method with a strong spatial overlap with reported templates [42]. Although other peer algorithms can detect five meta-FCs at their second layer, some reported FCs are different from templates. Specifically, *DICA* failed to detect the activated occipital lobe compared with templates presented in the first column, in Figure 3; in addition, there are some differences between extracted meta-FCs and templates, e.g., FCs in the first and second column; in detail, the template FCs in first column contains the occipital lobe that are disrupted in most meta-FCs identified via deep linear learning method; furthermore, the FCs identified by *SDDL* in third and fourth column presents a disruption of areas in activation areas of canonical Executive Control Network; similarly, *DBN* can only perfectly reveal the meta-FCs in the last column.

Moreover, the quantitative results are included in Figure 5(c) to compare the similarity between identified meta-FCs with corresponding templates. We further provide theoretical explanations of why *SENDER* can provide more FCs/spatial features than *DICA* in the proof in Theorem 2.2, Appendix B, Supplementary Material.

Note that another contribution of SENDER is to provide a variable including the potentially corresponding sub-FCs, while other peer algorithms cannot build the relations between the variables in method and sub-FCs using the same hyperparameters. For instance, in the third column of Figure 4, a precuneus, a functional core of Default Mode Network (DMN), is detected. However, these minor/sub-FCs could be more sensitive to some brain diseases from a clinical translational perspective that can guide the personalized diagnosis and treatment [35, 36]. Moreover, as discussed before, these sub-FCs are designed clearly as variables $\{S_i\}_{i=1}^{k}$ of *SENDER* introduced in Eq. (1) rather than randomly extracted features [5].

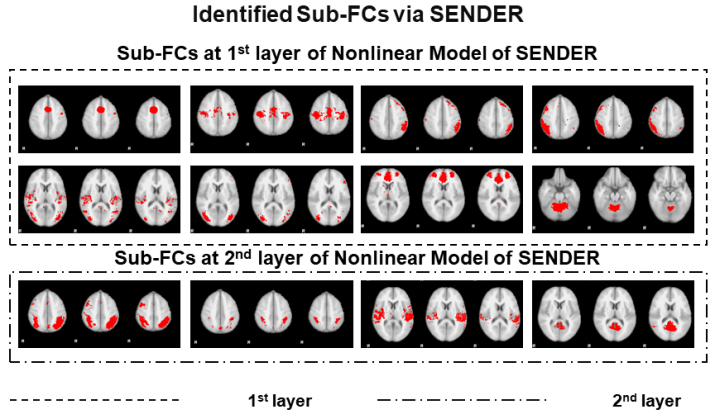

Figure 4. The qualitative comparison of sub-FCs identified by SENDER. These sub-FCs are derived via shallow and deeper layers of the background component matrix of SENDER.

In short, Figure 5 (a) indicates the potential organizations of canonical, meta, and sub-FCs. To validate the reconstruction performance of SENDER and the other three peer methods, we calculate the similarity of all FCs extracted by SENDER and the other four peer methods; SENDER shows an overall higher similarity than the four peer algorithms.

Moreover, since there has not been a rigorous 'ground-truth' to quantitatively validate the performance of SENDER and the other four peer methods, especially for meta-FCs, we alternatively investigate the identifiability [40] of canonical and meta-FCs. Calculating the identifiability can further validate the consistency and reproducibility of SENDER and other four peer methods in a data-driven fashion. At first, we randomly separate the original input data into two independent sets as $FC_{test}$ and $FC_{retest}$ shown in Eq. (9-1), and calculate the identifiability using Eq (9-2). The quantitative identifiability of meta-FCs is shown in Figures 6 (a) and (b). As discussed, the following equations detail the procedure of calculating identifiability:

$$fc_i \in FC_{test}, fc_j \in FC_{retest}, FC_{test} \cap FC_{retest} = \emptyset \quad (9\text{-}1)$$

$$identifiability \leftarrow \frac{\sum_i \sum_j corr(fc_i, fc_j)}{|FC_{test}| \times |FC_{retest}|} \quad (9\text{-}2)$$

In Eqs. (9-1) and (9-2), $corr(fc_i, fc_j)$ represents the calculation of the corresponding components identified from test and retest data using Pearson Correlation [40]. In this work, the calculation of correlation is replaced by Intraclass Correlation Coefficient (ICC). And $|FC_{test}|$, $|FC_{retest}|$ represent the number of components in the test and retest datasets, respectively.

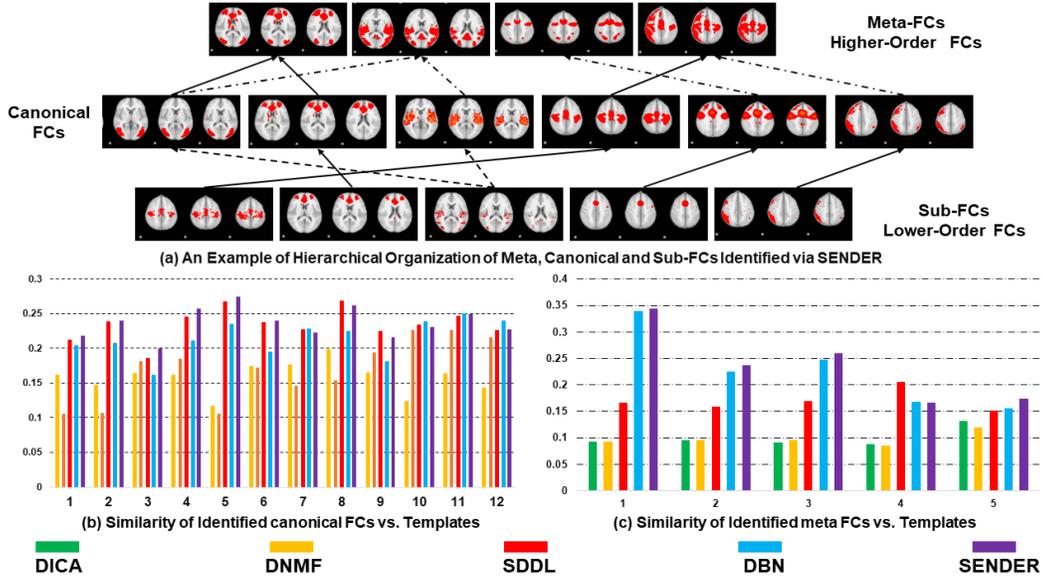

Figure 5. (a) An example of hierarchical structure on FCs presents the hierarchical organizations identified via linear/nonlinear methods and background matrix in SENDER. The bottom slices represent the sub-FCs which include partial/minor functional regions of canonical FCs. The middle slices show the identified canonical FCs and the top slices provide the meta FCs extracted via SENDER. The dashed line indicates the high-order FCs do not entirely include lower-order FCs and other regions are involved. (b) and (c) provide the similarity of canonical and meta-FCs derived by SENDER with the templates, shown in the first row in Figures 1, 2, and 3, respectively.

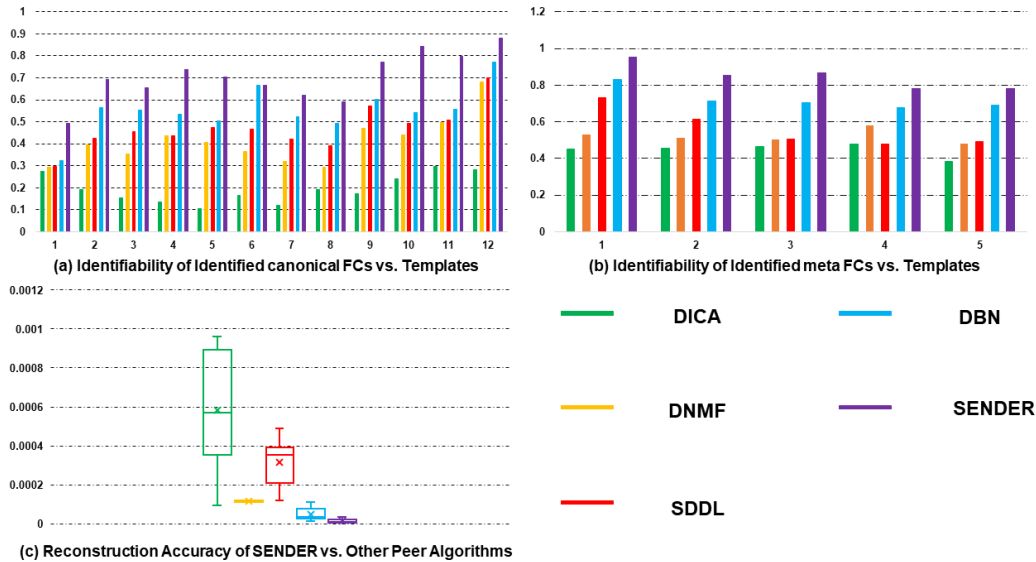

Figure 6. (a) and (b) show the identifiability comparison of identified canonical and meta-FCs via SENDER and the other three peer methods based on all subjects' resting-state fMRI signals from healthy individuals in CNP (https://openfmri.org/dataset/ds000030/). And (c) provides the comparison of reconstruction accuracy, i.e., training loss, of SENDER and the other four peer methods in the second layer.

In Figures 6 (a) and (b), the quantitative results of identifiability demonstrate that the identified canonical and meta-FCs via linear/nonlinear methods in SENDER provide higher identifiable over the other four peer methods [40]. In addition, in Figure 6 (c), the significant difference in reconstruction accuracy can be easily observed since the purple box plot shows the highest accuracy of training loss at the second layer.

## 4    Conclusion

In this work, the proposed SENDER adopts STORM optimizer, the alternative optimization strategy, and *RRO*, for data-driven determination of all hyperparameters, to reveal the canonical, meta, and sub-FCs. Furthermore, the hybrid modeling and efficient non-fully connected architectures of SENDER enable the discovery of canonical, meta, and sub-FCs with the highest identifiability to the other four peer methods. Moreover, the results show that the meta-FCs can even be detected in shallow layers. Finally, the theoretical studies and experimental validation further indicate an accurate approximation to original input and high convergence rate of SENDER, which are comparable to or even better than other peer algorithms, such as DNNs.

Moreover, SENDER can potentially synergize research of neurodevelopmental, neurodegenerative, and psychiatric disorders since these revealed novel canonical, meta, and sub-FCs that can be generated as the clinical biomarkers benefitting personalized diagnosis, prognosis and treatment monitoring [35-39].

Overall, we believe that SENDER can play a role as an inspiring deep hybrid learning method for a fruitful future with a profound influence on facilitating the research of deep learning methods, computational neuroscience, and clinical translational application.

# Appendix A

The matrix polynomials are defined as: $\forall k \in \mathbb{N}\ P_{2k}(X) = (XX^T)^k$, $P_{2k+1}(X) = (XX^T)^k X$, $X \in \mathbb{R}^{S \times T}$; and $\{P_n(X)\}_{n=1}^{N}$ defines a series of matrix polynomials, for example: $\{P_n(X)\}_{n=1}^{3} = \{X, XX^T, XX^T X\}$, $X \in \mathbb{R}^{S \times T}$;

Moreover, it is easy to prove $\{P_n\}_{n=1}^{\infty}$ denoted on $\mathbb{R}^{S \times T}$ as a ring $(\{P_n\}_{n=1}^{\infty}, +, \times)$, and it also demonstrates $\{P_n(X)\}_{n=1}^{N} \in (\{P_n\}_{n=1}^{\infty}, +, \times)$, e.g., $(\{P_n\}_{n=1}^{\infty}, +, \times) \supseteq \sum_{i=1}^{M}\{P_n(X)\}_{n=1}^{N}$. (Dummit, 2004; Kadison, 1997). Then we introduce the theorem 1.1 to describe the superiority of DEMAND.

***Theorem 1.1*** (Accurate Approximation of SENDER) Given a real function $f: \mathbb{R}^{P \times Q} \to \mathbb{R}^{P \times Q} \cup \{\pm\infty\}$ and $m(\{I \in \mathbb{R}^{M \times N}: f(I) = \pm\infty\}) = 0$ where $m(\cdot)$ represents the Lebesgue measure [29]. Given $I \in \mathbb{R}^{M \times N}$, SENDER includes a linear model and nonlinear model with multiple activation functions can be denoted as $\{P_k(I)\}_{k=1}^{N_1}$ and $\{\mathcal{N}_k(I)\}_{k=1}^{N_2}$ we have: if P denotes a series of matrix polynomials, such as, and $\mathcal{N}$ denotes a smooth activation function, and $\forall\ \varepsilon > 0$, $N > 0$, $N_1 > N$, $N_2 < \infty$, $\left\|\{P_k(I)\}_{k=1}^{N_1} + \{\mathcal{N}_k(I)\}_{k=1}^{N_2} - f(I)\right\| \leq \varepsilon$.

***Proof***: According to Лузин (Luzin) Theorem in [29], we have a close set:
$$F_n \subset F_{n+1} \subset \cdots \subseteq \mathbb{R}^{M \times N}$$

$$m(\mathbb{R}^{M \times N} \setminus F_k) = \frac{1}{k}, k \in \mathbb{N} \tag{A.1}$$

$$f \in C(F_k)$$

Then we have a consistent real function $g(X)$, and obviously we have:
$$g(X) = f(X) \tag{A.2}$$

Since for any continuous real function, we have:
$$|g(X) - P_k(X)| < \frac{1}{k} \tag{A.3}$$

Let $\mathcal{F} = \bigcup_{k=1}^{\infty} F_k$, and obviously we have:
$$m(\mathbb{R}^{M \times N} \setminus F_k) = m(\mathbb{R}^{M \times N} \setminus \bigcup_{k=1}^{\infty} F_k) = \bigcap_{k=1}^{\infty} m(\mathbb{R}^{S \times T} \setminus F_k) = \bigcap_{k=1}^{\infty} \frac{1}{k} = 0 \tag{A.4}$$

Moreover, it is easy to prove $\{P_n\}_{n=1}^{\infty}$ denoted on $\mathbb{R}^{S \times T}$ as a ring $(\{P_n\}_{n=1}^{\infty}, +, \times)$, and it also demonstrates $\{P_n(X)\}_{n=1}^{N} \subseteq (\{P_n\}_{n=1}^{\infty}, +, \times)$, e.g., $P_n(X) \stackrel{\text{def}}{=} \prod_{i=1}^{N} x_i + \sum_{j=1}^{N} y_j$.

If $\mathfrak{F}$ is a real function denoted on set $\mathcal{F}$, it indicates:
$$\lim_{N \to \infty} |\mathfrak{F} - \{P_n(X)\}_{n=1}^{N}| = 0 \tag{A.5}$$

then we have $\lim_{N \to \infty} \{P_n(X)\}_{n=1}^{N} = \mathfrak{F}$, meanwhile, if $N$ is large enough, $|\mathfrak{F} - \{P_n(X)\}_{n=1}^{N}| < \varepsilon$ holds. Moreover, if $\{\mathcal{N}_n(X)\}_{n=1}^{N} \in C(\mathbb{R}^{S \times T})$, according to Theorem, $|\mathcal{N}_n(X) - \{P_n(X)\}_{n=1}^{\infty}| \to 0$; we have:
$$|\mathfrak{F} - \{\mathcal{N}_n(X)\}_{n=1}^{N}| = |\mathfrak{F} - \{P_n(X)\}_{n=1}^{\infty}| < \frac{\varepsilon}{2} \tag{A.6}$$

Thus, considering $\forall\ \varepsilon > 0$, $N > 0$, $N_1 > N$, $N_2 < \infty$, rewrite Eq. (6) as:
$$\left|\mathfrak{F} - \{P_n(X)\}_{n=1}^{N_1}\right| < \frac{\varepsilon}{2} \tag{A.7-1}$$

$$\left|\mathfrak{F} - \{\mathcal{N}_n(X)\}_{n=1}^{N_2}\right| < \frac{\varepsilon}{2} \tag{A.7-2}$$

Then, obviously, we have Eq. (A.8) hold as below:

$$|\mathfrak{F} - \{P_n(X)\}_{n=1}^{N_1} - \{\mathcal{N}_n(X)\}_{n=1}^{N_2}| < \varepsilon \tag{A.8}$$

***Definition 1.2* (Variance Bounded Real Function)** Given a real function $f$ denoted on $[a,b]$, and $\Delta: a = x_0 < x_1 < x_2 < \cdots < x_n = b$. A sum as $v_\Delta = \sum_{i=1}^{n}|f(x_i) - f(x_{i-1})|$ and $V_a^b(f) = \sup\{v_\Delta: \forall \Delta\}$. The variance bounded real function is denoted as $V_a^b(f) < \infty$.

***Definition 1.3* (Amplitude of Real Function)** Given a real function $f$ denoted on $[a,b]$, and $\forall B(x_0, \delta) \subseteq [a,b], \delta > 0$; $\omega_f(x_0) = \lim_{\delta \to 0} \sup\{|f(x') - f(x'')|: x', x'' \in B(x_0, \delta)\}$.

***Lemma 1.1* (Smooth & Variance Bounded Real Function)** If and only if a real function $f \in Lip1([a,b])$, $V_a^x(f) < \infty$ holds.

***Proof***: If $f \in Lip1([a,b])$, it indicates: $\forall x_1, x_2 \in [a,b]$, we have: $|f(x_1) - f(x_2)| \leq L|x_1 - x_2|$. Moreover, according to Definition 1, we have:

$$v_\Delta = \sum_{i=1}^{n}|f(x_i) - f(x_{i-1})| \leq L(|x_0 - x_1| + |x_1 - x_2| + \cdots + |x_n - x_{n-1}|) \tag{A.9}$$
$$\leq L(b-a) < \infty$$

If $V_a^x(f) < \infty$ holds, it demonstrates: $\sup\{v_\Delta: \forall \Delta\} < \infty$, let $x_n$ be $x$, we have:

$$\sum_{i=1}^{n}|f(x_i) - f(x_{i-1})| \leq \sup\{v_\Delta: \forall \Delta\} < \infty \tag{A.10}$$

Furthermore, if $n \to \infty$, and $\sum_{i=1}^{n}|f(x_i) - f(x_{i-1})| < \infty$ holds, obviously, we must have:
$$|f(x_i) - f(x_{i-1})| \to 0 \tag{A.11}$$
It should satisfy:
$$x_i, x_{i-1} \in B(x, \varepsilon), \forall \varepsilon > 0 \; |f(x_i) - f(x_{i-1})| \leq L|x_i - x_{i-1}| \tag{A.12}$$
Since $\forall x_i, x_{i-1} \in [a,b]$, obviously, we have:
$$f \in Lip1([a,b])$$

***Lemma 1.2* (Amplitude & Variance Bounded Real Function)** $\omega_f(x_0) = \lim_{\delta \to 0} \sup\{|f(x') - f(x'')|: x', x'' \in B(x_0, \delta) \subseteq [a,b]\} < \varepsilon$ is equivalent to $f \in Lip1([a,b])$.

***Proof***: If $f \in Lip1([a,b])$ holds, similarly, if $n \to \infty$, $\forall \{x_i\}_{i=1}^{n}$, we have
$$|f(x_i) - f(x_{i-1})| \to 0 \tag{A.13}$$
Replace $x_i$ and $x_{i-1}$ by $x', x''$, respectively, it satisfies:
$$\omega_f(x_0) = \lim_{\delta \to 0} \sup\{|f(x') - f(x'')|: x', x'' \in B(x_0, \delta) \subseteq [a,b]\} < M \tag{A.14}$$
If $\omega_f(x_0) = \lim_{\delta \to 0} \sup\{|f(x') - f(x'')|: x', x'' \in B(x_0, \delta) \subseteq [a,b]\} < \infty$, we assume:
$$\omega_f(x_0) = \lim_{\delta \to 0} \sup\{|f(x') - f(x'')|: x', x'' \in B(x_0, \delta) \subseteq [a,b]\} < \varepsilon \tag{A.15}$$
Obviously, given $x', x'' \in B(x_0, \delta) \subseteq [a,b]$, we have:
$$\forall \varepsilon > 0 \; |f(x') - f(x'')| < \varepsilon \tag{A.16}$$
When $\delta \to 0$, let $|x' - x''| = \frac{\varepsilon}{L}$, it also indicates:
$$|f(x') - f(x'')| < L|x' - x''| \tag{A.17}$$

***Lemma 1.3* (Cantor Theorem)** Given $\{B_i\}_{i=1}^{\infty}$ are closed sets and $\forall B_i \neq \emptyset$, if $B_1 \supseteq B_2 \supseteq \cdots \supseteq B_k \supseteq \cdots$, $\cap_{i=1}^{\infty} B_i \neq \emptyset$.

***Lemma 1.4* (Vitali Covering Lemma)** Given $\{B_i\}_{i=1}^{n}$ are closed sets and $\forall B_i \cap B_j = \emptyset, i \neq j, E \subseteq \mathbb{R}$, and $m^*(E) < \infty$, if $m^*(E \setminus \cup_{i=1}^{n} B_i) < \varepsilon, \forall \varepsilon > 0$, holds, $\{B_i\}_{i=1}^{n}$ defines a Vitali Covering of $E$.

***Lemma 1.5*** **(Heine-Borel Covering Theorem)** Given $\Gamma$ is a close and bounded set. Then, an open set sequence as $\{g_i\}_{i=1}^K \overset{\text{def}}{=} G$, $\cup_{i=1}^K g_i \supseteq \Gamma$, and $\bar{\bar{G}} = \aleph_0$.

***Lemma 1.6*** **(Composition of Function)** Given $g \in Lip1([a,b])$, and $g$ is not a constant real function, if $f \notin Lip1([a,b])$, $g(f(x)) \notin Lip1([a,b])$ holds.

**Proof**: Proof by contradiction, if assume $g(f(x)) \in Lip1([a,b])$, $\forall x_1, x_2 \in [a,b]$, $f(x_1), f(x_2) \in [a,b]$,

$$|g(f(x_1)) - g(f(x_2))| < L_g|f(x_1) - f(x_2)| < N < \infty \text{ and } L_g \neq 0. \quad (A.18)$$

However, since $f \notin Lip1([a,b])$, we have: $|f(x_1) - f(x_2)| > M$ that is contradiction with $|f(x_1) - f(x_2)| < \frac{N}{L_g}$. Thus, $(f(x)) \notin Lip1([a,b])$ holds.

***Theorem 1.2*** **(Efficiency of Non-Fully Connected Architecture of SENDER)** Given a series of non-smoothed activation function $\{f_i\}_{i=1}^N$, denoted on $[a,b] \subseteq \mathbb{R}^1$, if assume $\{f_i\}_{i=1_i}^N \subseteq Lip1([a,b]\setminus\{U(x_i, \delta)\})$ $i \in \mathbb{N}$, and $U(x_i, \delta)$ is an open cubic with the center $x_i$ and radius $\delta > 0$. The composition of $f_i, f_j \in \{f_i\}_{i=1}^N$ are denoted as $f_{j,i} \overset{\text{def}}{=} f_j(f_i(x))$; the various composition $\mathcal{F} \overset{\text{def}}{=} f_{\cdots,k,\cdots j,i} \in Lip1([a,b]\setminus[c,d])$ holds, when $k \to \infty$, $[c,d] \supseteq \cup_{i=1}^k U(x_i, \delta)$ and $m([c,d]) \neq 0$; moreover, given $t \to \infty$, the summation of $\sum_{i=1}^t \mathcal{F}_i$ leads to $\sum_{i=1}^t \mathcal{F}_i \notin Lip1([a,b]\setminus[c',d'])$, $[c',d'] \supseteq [c,d]$, and $m([c',d']) \neq 0$. And $m(\cdot)$ represents the Lebesgue measure.

**Proof**: At first, we discuss $k < \infty$, and we assume, $f \in Lip1([a,b]\setminus\{x_0\})$

According to Lemma 1 and Lemma 2, if $x_0 \in B(x_0, \delta)$, for we have:

$$\omega_{f_i}(x_i) = \lim_{\delta \to 0} \sup\{|f_i(x') - f_i(x'')| : x', x'' \in B(x_i, \delta) \subseteq [a,b]\} > M \quad (A.19)$$

$$\omega_{f_j}(x_j) = \lim_{\delta \to 0} \sup\{|f_j(y') - f_j(y'')| : y', y'' \in B(x_j, \delta) \subseteq [a,b]\} > M \quad (A.20)$$

Thus, we have $f_i$ and $f_j$ are not smooth on $B(x_i, \delta)$ and $B(x_j, \delta)$, respectively.

And for the composition, let $k = 2$,

$$\omega_{f_{j,i}}(x_i) = \lim_{\delta \to 0} \sup\{|f_j(f_i(x')) - f_j(f_i(x''))| : x', x'' \in B(x_i, \delta) \subseteq [a,b]\} \quad (A.21)$$

Let $f_i(x') = x'_j$ and $f_i(x'') = x''_j$, if $(x'_j, x''_j) \in B(x_k, \delta)$, it is easy to prove the amplitude of $f_{j,i}$ as following:

$$\omega_{f_{j,i}}(x_1) = \lim_{\delta \to 0} \sup\{|f_2(x'_j) - f_2(x''_j)| : x'_j, x''_j \in B(x_k, \delta) \subseteq B(x_j, \delta) \subseteq [a,b]\} \quad (A.22)$$
$$> M$$

Naturally, we need to analyze other relations of $B(x_k, \delta)$ and $B(x_j, \delta)$; in detail, there are five situations to be discussed separately:

1). Assume, if $\forall B(x_k, \delta) \cap B(x_j, \delta) = \emptyset, i, j \in \mathbb{R}, i \neq j$, obviously, due to the same composition, we have: $B(x_k, \delta) \cap B(x_{k-1}, \delta) = \emptyset$, according to Lemma 1.3,

$$[c,d] \setminus \bigcup_{k=1}^{\infty} B(x_k, \delta) = \{\hat{x}_j\}_{j=1}^N \quad (A.23)$$

And

$$m\left(\{\hat{x}_j\}_{j=1}^N\right) = 0 \quad (A.24)$$

According to Lemma 1.6, $f_j(f_i(B(x_i, \delta)) \notin Lip1(B(x_k, \delta) \cup B(x_j, \delta))$, therefore, we have:

$$f_{\cdots,k,\cdots j,i} \notin Lip1([c,d]\setminus\{\hat{x}_j\}_{j=1}^N) \quad (A.25)$$

2). Similarly, if we assume $\forall B(x_k, \delta) \cap B(x_j, \delta) \neq \emptyset$, due to the composition, we have: $B(x_k, \delta) \cap B(x_{k-1}, \delta) \neq \emptyset$, according to Lemma 1.5,

$$[a,b] \supseteq \bigcup_{k=1}^K B(x_k, \delta) \supseteq [c,d] \quad (A.26)$$

Therefore, we can conclude:
$$f_{\cdots,k,\cdots j.i} \notin Lip1([c,d]) \tag{A.27}$$

3). Moreover, if $B(x_j, \delta) \supseteq B(x_k, \delta) \supseteq B(x_{k+1}, \delta) \supseteq \cdots$, according to Lemma 1.5, we have:
$$\cap_{i=1}^{K} B(x_i, \delta) = \Xi \neq \emptyset \tag{A.28}$$

It means:
$$f_{\cdots,k,\cdots j.i} \in Lip1([a,b]\backslash B(x_j, \delta)) \tag{A.29}$$

Thus, similarly, we have:
$$f_{\cdots,k,\cdots j.i} \in Lip1([a,b]\backslash \Xi) \tag{A.31}$$

4). Finally, if $\cdots \supseteq B(x_{k+1}, \delta) \supseteq B(x_k, \delta) \supseteq B(x_j, \delta)$,

Therefore, based on (1) and (2), we have:
$$\omega_{f_{k,\cdots,2,1}}(x_2) = \lim_{\delta \to 0} \sup\{|f_{k,\cdots,2,1}(x') - f_{k,\cdots,2,1}(x'')|: x', x'' \in [c,d]\} > M \tag{A.32}$$

It indicates:
$$f_{\cdots,k,\cdots j.i} \in Lip1([a,b]\backslash \bigcup_{i=1}^{\infty} B(x_i, \delta)) \tag{A.33}$$

5). Comprehensively, the situation includes all previously discussed (1) to (4), it is easy to conclude:
$$f_{\cdots,k,\cdots j.i} \in Lip1([a,b]\backslash [c,d] \tag{A.34}$$

Using Lemma 1.1, obviously, given $\Delta: a = x_0 < x_1 < x_2 < \cdots < x_n = b$, and $\Delta': x' < \hat{x}_1 < \hat{x}_2 < \cdots < \hat{x}_n < x''$, $v_{\Delta_1} + v_{\Delta_2} = v_\Delta$.

$$v_\Delta + v_{\Delta'} = v_{\Delta_1} + v_{\Delta_2} + v_{\Delta'}$$
$$= \sum_{i=1}^{n_1} |f(x_i) - f(x_{i-1})| + \sum_{i=n_1}^{n_2} |f(x_i) - f(x_{i-1})| + \sum_{i=n_2}^{n} |f(x_i) - f(x_{i-1})| \tag{A.35}$$

Since $v_{\Delta'} > M$, $v_\Delta + v_{\Delta'} > M$, it is easy to have:
$$\sum_{i=1}^{t} f_i^t \notin Lip1([a,b]) \tag{A.36}$$

***Lemma 1.7*** (Contraction of STORM Operator) If denote $STORM \stackrel{\text{def}}{=} \mathcal{T}: \mathbb{R}^{P \times Q} \to \mathbb{R}^{P \times Q}$, operator $\mathcal{T}$ is a bounded contraction operator.

***Proof***: According to definition of contraction operator, we have:
$$\|\mathcal{T}^{t+k}X - \mathcal{T}^t X\| = \left\|\sum_{i=t+1}^{t+k} \eta_t d_t\right\| \tag{A.37}$$

Considering the equivalence of norms in finite dimensional space and Cauchy-Schwarz inequality, we have:
$$\left\|\sum_{i=t+1}^{t+k} \eta_t d_t\right\|_2^2 \leq \left\|\sum_{i=t+1}^{t+k} \eta_t^2\right\|_2^2 \cdot \left\|\sum_{i=t+1}^{t+k} d_t^2\right\|_2^2 \tag{A.38}$$

In [27], due to the definition of $\eta_t$:
$$\eta_t \stackrel{\text{def}}{=} \frac{k}{(\omega + \sum_{i=1}^{t} G_t)^{1/3}} \tag{A.39}$$

According to the definition of $G_t$ as $\|\nabla f(x_t, \xi_t)\|$, if we assume the target function is smooth and variance bouded, we have:

$$\|\nabla f(x_t, \xi_t)\| \leq M$$

$$|\eta_t| < \varepsilon \tag{A.40}$$

Then, given $t > N$, $N \in \mathbb{N}$, we can derive the following formula:

$$\left\|\sum_{i=t+1}^{t+k} \eta_t d_t\right\|_2^2 \leq \varepsilon \cdot \left\|\sum_{i=t+1}^{t+k} d_t^2\right\|_2^2 \tag{A.41}$$

In Eq (A.41), we only consider $\left\|\sum_{i=t+1}^{t+k} d_t^2\right\|$:

$$\|d_{t+2}X - d_{t+1}X\| \tag{A.42}$$
$$= \|\nabla f(x_{t+2}, \xi_{t+2}) - \nabla f(x_{t+1}, \xi_{t+1}) + (1 - \alpha_{t+2})(d_{t+1} - \nabla f(x_{t+1}, \xi_{t+2}) - (1 - \alpha_{t+1})(d_t - \nabla f(x_t, \xi_{t+1})\|$$

We can easily conclude:

$$\|d_{t+2}X - d_{t+1}X\| \tag{A.43}$$
$$\leq \|\nabla f(x_{t+2}, \xi_{t+2}) - \nabla f(x_{t+1}, \xi_{t+1})\|$$
$$+ \|(1 - \alpha_{t+2})(d_{t+1} - \nabla f(x_{t+1}, \xi_{t+2}) - (1 - \alpha_{t+1})(d_t - \nabla f(x_t, \xi_{t+1})\|$$

Then we have:

$$\|d_{t+2}X - d_{t+1}X\|$$
$$\leq \|\nabla f(x_{t+2}, \xi_{t+2}) - \nabla f(x_{t+1}, \xi_{t+1})\|$$
$$+ \|d_{t+1} - d_t + \alpha_{t+1}d_t - \alpha_{t+2}d_{t+1}\| \tag{A.44}$$
$$+ \|\nabla f(x_{t+1}, \xi_{t+2}) - \nabla f(x_t, \xi_{t+1})\|$$

And we can also conclude:

$$\|d_{t+2}X - d_{t+1}X\| \leq \varepsilon_1 + \|d_{t+1} - d_t + \alpha_{t+1}d_t - \alpha_{t+2}d_{t+1}\| + \varepsilon_2 \tag{A.45}$$

Eq. (A.45) can be rewritten as below:

$$\|d_{t+2}X - d_{t+1}X\| \leq \varepsilon_1 + (1 - \hat{\alpha}) \cdot \|d_{t+1} - d_t\| + \varepsilon_2 \tag{A.46}$$
$$\hat{\alpha} \stackrel{\text{def}}{=} \min(\alpha_{t+1}, \alpha_{t+2})$$

Obviously, $\|d_{t+1} - d_t\| \leq M$, we have:

$$\frac{\|d_{t+2}X - d_{t+1}X\|}{\|d_{t+1} - d_t\|} \leq (1 - \hat{\alpha}) + \varepsilon_1 + \varepsilon_2 \tag{A.47}$$

If $c > 0$, according to the definition of $\alpha$:

$$\alpha_{t+1} \stackrel{\text{def}}{=} c \cdot \eta_t^2 \tag{A.48}$$

Then, due to $\forall \varepsilon_1, \varepsilon_2$, Eq. (A.47) holds:

$$\frac{\|d_{t+2}X - d_{t+1}X\|}{\|d_{t+1} - d_t\|} \leq (1 - \hat{\alpha}) \tag{A.49}$$

Since $\forall t \in \mathbb{N}, \alpha_t > 0$:

$$\frac{\|d_{t+2}X - d_{t+1}X\|}{\|d_{t+1} - d_t\|} < 1 \tag{A.50}$$

Thus, we proved the STORM operator $\mathcal{T}$ is a contraction operator within the finite dimensional space.

***Lemma 1.8* (Contraction of Operators Combination)** Given two contraction mappings $\Phi_1$ and $\Phi_2$, we have the composite of two contraction mapping as $\Phi_2 \cdot \Phi_1$. The composite mapping $\Phi_2 \cdot \Phi_1$ must be contractive.

**Proof**: According to the definition of contraction linear operator, we have:

$$\exists \zeta \in (0,1)$$
$$\rho \stackrel{\text{def}}{=} \|\Phi x - \Phi y\| \tag{A.51}$$
$$\rho(\Phi x, \Phi y) \leq \zeta \rho(x, y)$$

Obviously, and we have:

$$\rho(\Phi_1 u, \Phi_1 v) \leq \zeta \rho(u, v) \ \forall \zeta \in (0,1) \tag{A.52}$$
$$\rho(\Phi_2 x, \Phi_2 y) \leq \eta \rho(x, y) \ \forall \eta \in (0,1)$$

If we set:

$$x = \Phi_1 u, y = \Phi_1 v \tag{A.53}$$

the inequality below holds:

$$\rho(\Phi_2 x, \Phi_2 y) \leq \eta \rho(\Phi_1 u, \Phi_1 v) \leq \zeta \eta \rho(u, v) \tag{A.54}$$

Since the definition as

$$\forall \zeta, \eta \in (0,1), \rho(\Phi_2 \Phi_1 u, \Phi_2 \Phi_1 y) \leq \zeta \eta \rho(u, v) \tag{A.55}$$

***Theorem 1.3* (Convergence Rate of SENDER in Finite Dimensionality Space)** Denote Adam as an operator $\mathcal{T}$ in a finite dimensionality space. Due to the convergence of STORM with a rate of $\mathcal{O}(T^{\frac{1}{2}} + \sigma^{\frac{1}{3}}/T^{\frac{1}{3}})$, the convergence of SENDER is guaranteed to be the same as STORM.

***Lemma 1.8* (Adam Operator is bounded)** [27] If we denote the Adam optimizer operator as $\mathcal{T}: \mathbb{R}^{P \times Q} \to \mathbb{R}^{P \times Q}$, we have $\|\mathcal{T}\| < \frac{1}{\sqrt{T}}$.

***Corollary 1.1* (General Contraction Operator)** According to Lemma 1.2, if denote the operators $\{\Phi_i\}_{i=1}^{K}$, $\forall \Phi_i \ i \in \mathbb{N}$, $\Phi_i: \mathbb{R}^{S \times T} \to \mathbb{R}^{S \times T}$; considering any combination of operators: $\Phi_K \cdot \cdots \cdot \Phi_2 \cdot \Phi_1$, if at least a single operator $\Phi_i$ is contraction operator, and other operators are bounded, such as $\forall i \neq k \ \|\Phi_i\| \leq M$. If and only if $\prod_{i=1}^{K}\|\Phi_i\| < 1$, the combination of operator series $\Phi_K \cdot \cdots \cdot \Phi_2 \cdot \Phi_1$ is a contraction operator.

**Proof**: Obviously, according to Lemma 1.2, use a series as $\{\zeta_i\}_{i=1}^{K}$ to replace $\zeta, \eta \in (0,1)$,
Obviously, we have:
$$\zeta_i \in (0,1) \ i \in \mathbb{N} \tag{A.56}$$
$$\rho(\Phi_K \cdot \cdots \cdot \Phi_2 \Phi_1 u, \Phi_K \cdot \cdots \cdot \Phi_2 \Phi_1 y) \leq \zeta_K \cdot \cdots \zeta_2 \cdot \zeta_1 \cdot \rho(u, v)$$

Since $\zeta_K \cdot \cdots \zeta_2 \cdot \zeta_1 < 1$, we have proved this corollary.

***Corollary 1.2* (Iterative Contraction Operator)** According to Lemma 1.2, if denote the operators $\{\Phi_i\}_{i=1}^{K}$, $\forall \Phi_i \ i \in \mathbb{N}$, $\Phi_i: \mathbb{R}^{P \times Q} \to \mathbb{R}^{P \times Q}$; considering any combination of operators: $\Phi_K \cdot \cdots \cdot \Phi_2 \cdot \Phi_1$, if at least a single operator $\Phi_i$ is contraction operator, and other operators are bounded, such as $\forall i \neq k, \|\Phi_i\| \leq M$. If and only if $\lim_{n \to \infty} \prod_{i=1}^{K}\|\Phi_i\|^n = c < 1$, the combination of operator series $\Phi_K^n \cdot \cdots \cdot \Phi_2^n \cdot \Phi_1^n$.

**Proof**: Obviously, according to Lemma 3.1 and Corollary 3.1, use a series as $\{\zeta_i\}_{i=1}^{K}$ to replace $\zeta, \eta \in (0,1)$,
And we have:
$$\forall \zeta_i \in (0,1) \ i \in \mathbb{N} \tag{A.57}$$

$$\rho(\Phi_K^n \cdot \cdots \cdot \Phi_2^n \cdot \Phi_1^n u, \Phi_K^n \cdot \cdots \cdot \Phi_2^n \cdot \Phi_1^n y) < \zeta_i^n \cdot \cdots \cdot \zeta_2^n \cdot \zeta_1^n \cdot \rho(u,v)$$
Since $0 < \zeta_i^n \cdot \cdots \cdot \zeta_2^n \cdot \zeta_1^n < 1$, we have proved this corollary.

***Theorem 1.3*** **(Convergence of SENDER in Finite Dimensionality Space)** SENDER can converge as fast as STORM.

***Proof***: If we have $U, V \in \mathbb{R}^{m \times n}$, according to Theorems 3.2-3.4, and Lemma 3.2, the SENDER can be represented as
$$SENDER \stackrel{\text{def}}{=} (\mathcal{SRNJ})^k \cdot I : \mathbb{R}^{M \times N} \to \mathbb{R}^{M \times N} \tag{A.58}$$
According to Lemma 1.2, Corollary 1.1 and 1.2, we conclude:
$$\|\mathcal{T}^k U - \mathcal{T}^k V\| \le \rho^k \|U - V\| \tag{A.60}$$
and $0 < \rho^k < 1$ holds and $\rho^k$ equals to $\mathcal{O}(T^{\frac{1}{2}} + \sigma^{\frac{1}{3}}/T^{\frac{1}{3}})$ [27],

And given other definitions of operators adopted by DEMAND, obviously, since all norms of these operators are bounded, the following norm inequality holds:
$$\|(\mathcal{SRNA})^k \mathcal{J} \cdot U - (\mathcal{SRNA})^k \mathcal{J} \cdot V\| \le (\frac{N}{asr})^k \cdot \|U - V\| \tag{A.61}$$
If $0 < (\frac{N}{asr})^k < 1, k \to \infty$, can guarantee the convergence of SENDER comparable to STORM. It demonstrates that the convergence rate of SENDER would be equal to STORM, since the convergence rate of STORM has been proved as $\mathcal{O}(T^{\frac{1}{2}} + \sigma^{\frac{1}{3}}/T^{\frac{1}{3}})$, and $(\frac{N}{asr})^k$ can be rewritten as $\frac{1}{\sqrt{asr}}(\frac{N}{asr})^k \sqrt{asr}$, if and only if $(\frac{N}{asr})^k \sqrt{asr}$ is a constant $\mathcal{C}$, and let $\frac{1}{\sqrt{asr}}$ be $\mathcal{O}(T^{\frac{1}{2}} + \sigma^{\frac{1}{3}}/T^{\frac{1}{3}})$.

***Collaroy 1.3*** **(Convergence of SENDER in Infinite Dimensionality Space)** Given the infinite dimensionality space, the SENDER is denoted as an operator as $\mathcal{D}: \mathbb{R}^{\infty \times \infty} \to \mathbb{R}^{m \times n}$. If we assume $\mathcal{D}$ and $\mathbb{R}^{\infty \times \infty}$ can be defined as infinite matrix and each element can be represented as $d_{i,j} \in \mathcal{D}$ and $r_{i,j} \in \mathbb{R}^{\infty \times \infty}$, $i, j \to \infty$, respectively. $\mathcal{D}$ can converge, if and only if $d_{i,j}^k \cdot r_{i,j}$ should be $\mathcal{O}(\frac{1}{n^p})$.

$$D = \begin{bmatrix} d_1 \\ d_2 \\ \vdots \\ d_{n-1} \\ \vdots \end{bmatrix} \tag{A.62}$$

In Eq. (C.21), operator $D$ denotes an infinite dimensionality operator.

$$X = \begin{bmatrix} r_1 \\ r_2 \\ \vdots \\ r_{n-1} \\ \vdots \end{bmatrix} \tag{A.63}$$

In Eq. (C.21), without generality, input $X$ denotes an infinite dimensionality matrix.

Then, given the operator $D$ applied on input matrix $X$ as:
$$D \otimes X = \begin{bmatrix} d_1 r_1 \\ d_2 r_2 \\ \vdots \\ d_{n-1} r_{n-1} \\ \vdots \end{bmatrix} \tag{A.64}$$

Obviously, due to the inequality of norms in the infinite dimensionality space, we examine the $\ell_2$ norm as an example:
$$\|D \otimes X\|_2 = \sum_{i=1}^{\infty} (d_i r_i)^2 \tag{A.65}$$

Easily, we can conclude:

$$\|D^k \otimes X\|_2 = \sqrt{\sum_{i=1}^{\infty}(d_i^k r_i)^2} < \infty \Leftrightarrow \lim_{k \to \infty}(d_i^k r_i)^2 = \frac{1}{n^p} \; p > 1 \tag{A.66}$$

Therefore, we have proved the DEMAND converges in an infinite dimensionality space, if and only if each element of $D \cdot X$ as $\frac{1}{n^p}$, and $p > 1, n \in \mathbb{R}$.

***Lemma 1.9*** **(Convergence of Alternative Optimization of Real Function)** For a series of real function as $\{f_{i,j}\}_{i,j=1}^{\infty}$. If we have: $\lim_{i \to \infty} f_{i,j}(x) \to h_{M,j}, a.e. \, x \in [a,b]$ and $\lim_{j \to \infty} h_{M,j} \to g_{M,N}, a.e. \, x \in [a,b]$. Then, $\exists \lim_{k \to \infty} f_{i_k, j_k} \to g_{M,N} \, a.e. \, x \in [a,b]$ holds.

***Proof***: If considering the uniform convergence of $\{f_{i,j}\}_{i,j=1}^{\infty}$, since $\lim_{i \to \infty} f_{i,j} \to h_{M,j}, a.e. \, x \in [0,1]$ and $\lim_{j \to \infty} h_{M,j} \to g_{M,N}, a.e. \, x \in [0,1]$, according to Riez Theorem, $\exists \{f_{i_k, j_k}\}_{k=1}^{\infty}$.

$$|f_{i_k, j_k} - h_j| < \frac{\varepsilon}{2} \tag{A.67}$$

And we have:

$$|h_j - g| < \frac{\varepsilon}{2} \tag{A.68}$$

Then, we have:

$$|f_{i_k, j_k} - h_j| + |h_j - g| = |f_{i_k, j_k} - g| < \varepsilon \tag{A.69}$$

***Theorem 1.4*** **(Alternative Convergence of SENDER)** Given $\{\mathcal{F}_{i,j,k,t}\}_{i,j,k,t=1}^{\infty}$, $\{\mathcal{C}_{j,k,t}\}_{j,k,t=1}^{\infty}$, $\{\mathcal{H}_{k,t}\}_{k,t=1}^{\infty}$, and $\{\mathcal{K}_t\}_{t=1}^{\infty}$, are series of continuous operator [30] applied on a finite dimensional space, the series of operators. And we have: $\mathcal{F}_{i,j,k,t}, \mathcal{C}_{j,k,t}, \mathcal{H}_{k,t}, \mathcal{K}_t: \mathbb{R}^{P \times Q} \to \mathbb{R}^{P \times Q}$. If we have: $\lim_{i \to \infty} \mathcal{F}_{i,j} \to \mathcal{C}_{j,k,t}$, $\lim_{j \to \infty} \mathcal{C}_{j,k,t} \to \mathcal{H}_{k,t}$, $\lim_{k \to \infty} \mathcal{H}_{k,t} \to \mathcal{K}_t$, and $\lim_{t \to \infty} \mathcal{K}_t \to \mathcal{G}$. Then, $\exists \lim_{n \to \infty} \mathcal{F}_{i_n, j_n, k_n, t_n} \to \mathcal{G}$ holds.

***Proof***: According to Lemma 1.5, similarly, let constant $T < \infty$, we have:

$$\|\mathcal{F}_{i_n,j_n,k_n,t_n} - \mathcal{C}_{j_n,k_n,t_n}\| < \frac{\varepsilon}{4T} \tag{A.70}$$

Similarly, we have:

$$\|\mathcal{C}_{j_n,k_n,t_n} - \mathcal{H}_{k_n,t_n}\| < \frac{\varepsilon}{4T}$$
$$\|\mathcal{H}_{k_n,t_n} - \mathcal{K}_{t_n}\| < \frac{\varepsilon}{4T} \tag{A.71}$$
$$\|\mathcal{K}_{t_n} - \mathcal{G}\| < \frac{\varepsilon}{4T}$$

The following inequality holds:

$$\|\mathcal{F}_{i_k,j_k} - \mathcal{G}\| = \|\mathcal{F}_{i_n,j_n,k_n,t_n} - \mathcal{C}_{j_n,k_n,t_n}\| + \|\mathcal{C}_{j_n,k_n,t_n} - \mathcal{H}_{k_n,t_n}\| \tag{A.72}$$
$$+ \|\mathcal{H}_{k_n,t_n} - \mathcal{K}_{t_n}\| + \|\mathcal{K}_{t_n} - \mathcal{G}\| \leq \frac{\varepsilon}{T}$$

And we also have:

$$\|\mathcal{F}_{i_n,j_n,k_n,t_n}X - \mathcal{G}X\| \leq \|\mathcal{F}_{i_n,j_n,k_n,t_n} - \mathcal{G}\| \cdot \|X\| < T \cdot \frac{\varepsilon}{T} = \varepsilon \tag{A.73}$$

This equation indicates the operator can converge to a fixed point defined on Banach space, using alternative strategy and Banach Fixed Point Theorem (Rudin, 1973), if and only if $\lim_{k \to \infty} \|\mathcal{F}_{i_n,j_n,k_n,t_n}\| < 1$.

# Appendix B

**Algorithm 2.1 (Core Algorithm):** Semi-Estimated Nonlinear Deep Efficient Reconstructor (SENDER)

**Input:** $I \in \mathbb{R}^{P \times Q}$, $I$ is the input signal matrix; set $\lambda > 1$ as the penalty parameter; randomly initialize $\{X_k\}_{k=1}^M$, $\{Y_k\}_{k=1}^M$, $\{U_k\}_{k=1}^M$, $\{V_k\}_{k=1}^M$ and $\{S_k\}_{k=1}^M$;
Set $r$ as the initial estimated rank of $X_1, Y_1, U_1, V_1$ and layer $k$ as 0.
   **while** $\min(rank_Y, rank_V) > 1$
      update $X_k$ using Eq. (3-1);
      update $Y_k$ using Eq. (3-2);
      update $U_k$ using Eq. (3-3);
      update $V_k$ using Eq. (3-4);
      update $S_k$ using Eq. (3-5);
      use **Algorithm 2.2** to estimate $rank_Y$ of $Y_k$;
      use **Algorithm 2.2** to estimate $rank_V$ of $V_k$;
      $k \leftarrow k + 1$;
      $I \leftarrow I - S_k$;
  **end while**
  $M \leftarrow k$;
Use Algorithm 2.6 to perform matrix back propagation for $\{X_k\}_{k=1}^M$ and $\{Y_k\}_{k=1}^M$;
Use Algorithm 2.7 to perform matrix back propagation for, $\{U_k\}_{k=1}^M$ and $\{V_k\}_{k=1}^M$;
**Output:** $\{X_i\}_{i=1}^K \in \mathbb{R}^{M \times N}$, $\{Y_i\}_{i=1}^K \in \mathbb{R}^{M \times N}$, $\{U_i\}_{i=1}^K \in \mathbb{R}^{M \times N}$, $\{V_i\}_{i=1}^K \in \mathbb{R}^{M \times N}$, and $\{S_i\}_{i=1}^K \in \mathbb{R}^{M \times N}$;

**Algorithm 2.2:** Rank Reduction Operator (RRO)

**Input:** $Y_k \in \mathbb{R}^{n \times m}$, $Y_k$ is the feature matrix;
  $YR_k \leftarrow QR(Y_k)$;
  $minRank \leftarrow \min(1, size(Y_k))$;
  $estRank \leftarrow minRank - 1$;
  $diagYR_k \leftarrow abs(diag(YR_k))$;
  using **Algorithm 2.3** calculate the weighted difference of $diagYR_k$;
  set weighted difference of $diagYR_k$ as $wd$;
  $[rankMax1, posMax1] \leftarrow \max(wd)$;
  using **Algorithm 2.4** to calculate the weighted ratio of $diagYR_k$;
  set weighted ratio of $diagYR_k$ as $wr$;
  $[rankMax2, posMax2] \leftarrow \max(wr)$;
  **if** rankMax1 is equal to 1
     $estRank \leftarrow posMax1$;
  **end if**
  $valWR \leftarrow \text{find}(wr > rankMax2)$;
  **if** number (valWR) is equal to 1
     $estRank \leftarrow posMax1$;
  **end if**

using **Algorithm 2.5** to calculate the weighted correlation of $diagYR_k$;
  set weighted correlation of $diagYR_k$ as wc;
  $[rankMax3, posMax3] \leftarrow \max(wc)$;
  *if* rankMax1 is equal to 1
    $estRank \leftarrow posMax3$;
  *end if*
  $valWC \leftarrow$ find( *wc>rankMax3* );
  *if* number (valWC) is equal to 1
    $estRank \leftarrow posMax3$;
  *end if*
  $estRank \leftarrow \max(posMax1, posMax2, posMax3)$;
***Output:** estRank*

---

**Algorithm 2.3:** Weighted Difference (WD)

***Input**:* $Vec \in \mathbb{R}^{1 \times m}, Vec$ is a vector;
 *cumSum* $\leftarrow$ Calculate *cumulative sum of Vec;*
 *diffVec* $\leftarrow$ Calculate differences between adjacent elements of *Vec*;
 *resverseVec* $\leftarrow$ reverse *Vec*;
 $WD \leftarrow abs(diffVec) ./ reverseVec$;
***Output:** WD*

---

**Algorithm 2.4:** Weighted Ratio (WR)

***Input**:* $Vec \in \mathbb{R}^{1 \times m}, Vec$ is a vector;
  $L \leftarrow$ Calculate length of *Vec*;
 *ratioVec* $\leftarrow Vec(1:L-1) ./ Vec(2:L)$;
 $WR \leftarrow (L-2)*ratioVec ./ sum(ratioVec)$;
***Output:** WR*

---

**Algorithm 2.5:** Weighted Correlation (WC)

***Input**:* $Vec \in \mathbb{R}^{1 \times m}, Vec$ is a vector;
$WC \leftarrow$ Calculate the weight correlation using Eq. (7)
***Output:** WC*

---

**Algorithm 2.6: Matrix Back Propagation (MBP) for Linear Model**

***Input**:* $\{X_i\}_{i=1}^M \in \mathbb{R}^{T \times S}$, $\{Y_i\}_{i=1}^M \in \mathbb{R}^{T \times S}$, set $Z_i \leftarrow \prod_{j=1}^i X_i Y_i$;

 *for* $k = T$ to 1

    $\psi \leftarrow \prod_{i=1}^{k-1} X_i$;

    $D_k \leftarrow \psi^\dagger Z_k \hat{Y}_k^\dagger$;

    *if* $k<M$

      $\hat{\alpha}_k \leftarrow \alpha_M$

  *else*

    $\hat{\alpha}_k \leftarrow D_{k+1}\hat{\alpha}_{k+1}$,
  *end if*

  $\hat{Y}_k^+ \oplus \hat{Y}_k^- \leftarrow \hat{Y}_k$

  $Z_k \leftarrow I - \prod_{i=1}^k U_i \cdot (\mathcal{N}_k \cdot V_k)$

  $\hat{Y}_k^+ \leftarrow Y_k^+ \odot \sqrt{\frac{[\psi^T Z_k]^+ + [\psi^T\psi]^- \hat{Y}_k^+}{[\psi^T Z_k]^- + [\psi^T\psi]^+ \hat{Y}_k^+}};$

  $|\tilde{Y}_k^-| \leftarrow |\hat{Y}_k^-| \odot \sqrt{\frac{[\psi^T Z_k]^+ + [\psi^T\psi]^- |\tilde{Y}_k^-|}{[\psi^T Z_k]^- + [\psi^T\psi]^+ |\tilde{Y}_k^-|}};$

  $\tilde{Y}_k^- \leftarrow -|Y_k^-|;$

  $Y_k \leftarrow \tilde{Y}_k^+ \oplus \tilde{Y}_k^-;$

*end for*

**Output:** $\{X_i\}_{i=1}^M \in \mathbb{R}^{P\times Q}, \{Y_i\}_{i=1}^K \in \mathbb{R}^{P\times Q}$;

---

**Algorithm 2.7:** Matrix Backpropagation (MBP) for Nonlinear Model

**Input:** $\{U_k\}_{k=1}^M, \{V_k\}_{k=1}^M, I$, and set $E = 0.01$, MaxIter, $U_k = \lim_{it\to\infty} U_k^{it}$, $X_k = \lim_{it\to\infty} X_k^{it}$, and $Y_k = \lim_{it\to\infty} Y_k^{it}$, $s = U_k^{it} V_k^{it}$

**for** *it* in 1 to *MaxIter*

$$K \leftarrow \left(\prod_{k=1}^M U_k\right)^T \cdot \left(I - \prod_{k=1}^M X_k \cdot Y_M\right)$$

$$P_k^{it} \leftarrow (U_k^{it})^T U_k^{it} \cdot \prod_{k=1}^{M-1} \max(U_k)$$

$$c_k^{it} \leftarrow \max(U_{k-1}) \cdot \frac{d\mathcal{N}^{-1}(s)}{ds}, s = U_k^{it} V_k^{it}$$

$$d_k^{it} \leftarrow \frac{d\mathcal{N}^{-1}(s)}{ds}, s = U_k^{it} V_k^{it}$$

$$D_k^{it} \leftarrow (U_{k-1}^{it})^T \cdot \left(U_{k-1}^{it} \cdot \mathcal{N}^{-1}(s) - V_{k-1}^{it}\right) \odot d_k^{it} \cdot (V_k^{it})^T$$

$$V_k^{it+1} \leftarrow V_k^{it} - \frac{T}{2^{it}}(C_k^{it})$$

$$U_k^{it+1} \leftarrow U_k^{it} - \frac{T}{2^{it}}(D_k^{it})$$

*end*
**Output:** $\{U_k\}_{k=1}^M$ and $\{V_k\}_{k=1}^M$

---

***Theorem 2.1* (Rank Reduction Operator is bounded)** If we denote the sparse operator as $\mathcal{R}: \mathbb{R}^{S\times T} \to \mathbb{R}^{S\times T}$, we have $\|\mathcal{R}\| < \infty$.

**Proof**: According to the definition of operator norm (Rudin, 1973), $\|\mathcal{R}\| \leq sup \frac{\|\mathcal{R}X\|}{\|X\|}$; obviously, $\|\mathcal{R}X\|$ and $\|X\|$ is bounded, since both of norms are based on finite dimensional matrix. And if we denote:

$$X = \begin{bmatrix} a_1 \\ a_2 \\ \vdots \\ a_{n-1} \\ a_n \end{bmatrix}, \mathcal{R}X = \begin{bmatrix} a_1 \\ a_2 \\ \vdots \\ a_k \\ \vdots \\ a_{n-1} \end{bmatrix} \tag{B.1}$$

Eq. (C11) implies:

$$sup \frac{\|\mathcal{R}X\|}{\|X\|} = \frac{\sum_{i=1}^{n} a_i^2}{\sum_{i=u}^{p}(a_i - b_i)^2 + \sum_{i=v}^{q} a_i^2} < \infty \tag{B.2}$$

Also, if we examine the weighted ratio and weight difference, only considering the finite dimensional space, we have:

$$X = \begin{bmatrix} a_1 \\ a_2 \\ \vdots \\ a_{n-1} \\ a_n \end{bmatrix}, WR \cdot X = \begin{bmatrix} a_2/a_1 \\ a_3/a_2 \\ \vdots \\ a_k/a_{k-1} \\ \vdots \\ a_n/a_{n-1}\; 0 \end{bmatrix} WD \cdot X = \begin{bmatrix} a_2 - a_1 \\ a_3 - a_2 \\ \vdots \\ a_k - a_{k-1} \\ \vdots \\ a_n - a_{n-1} \end{bmatrix}$$

Obviously, for each rank estimation, the dimension of input matrix can be reduced at least by one. Similarly, $WR$, $WD$, and $WC$ can be considered as the contract operators for dimensional estimation. It demonstrates that the input matrix can be reduced to a vector by *n-1* iterations at most.

***Theorem 2.2*** **(Explanation of More Components Detected via SENDER Than ICA)** In general, DEMAND can detect more components from an input signal than Independent Component Analysis Method.

**Proof**: At first, we assume all component included in signal matrix as:

$$I \stackrel{\text{def}}{=} \bigcup_{i=1}^{M} \xi_i \subseteq \mathbb{R}^{T \times M} \tag{B.3}$$

A single component can be denoted as following:

$$\xi_i \stackrel{\text{def}}{=} [\xi_{1,i}, \xi_{2,i}, \cdots \xi_{T,i}] \tag{B.4}$$

And we assume that there is no any overlap in these components:

$$\forall i \neq j \; \xi_i \cap \xi_j = \emptyset \tag{B.5}$$

If we define ICA operator as below:

$$ICA \stackrel{\text{def}}{=} \mathcal{T}: \mathbb{R}^{T \times M} \rightarrow \mathbb{R}^{1 \times M} \tag{B.6}$$

Obviously, when ICA is applied on the input signal, it is easy to conclude:

$$\mathcal{T} \cdot I = \mathcal{T}\left(\bigcup_{i=1}^{M} \xi_i\right) = [\xi_1, \xi_2, \cdots, \xi_M] \tag{B.7}$$

Similarly, as previous definition of DEMAND as operator $\mathcal{D}$, we can have:

$$\mathcal{D} \cdot I = D\left(\bigcup_{i=1}^{M} \xi_i\right) \tag{B.8}$$
$$= [\xi_1, \xi_2, \cdots, \xi_M, (\xi_1, \xi_2), (\xi_1, \xi_3), \cdots, (\xi_1, \xi_3, \cdots, \xi_k), \cdots]$$

It is easy to calculate the number of components detected by ICA, due to independent constraint:
$$|\mathcal{T} \cdot I| = M \tag{B.9}$$

Nevertheless, we can conclude:
$$|\mathcal{D} \cdot I| = 2^M \tag{B.10}$$

Obviously, we also have:
$$M \ll 2^M \tag{B.11}$$

Inequality (B.9) demonstrates that the number of components identified by SENDER should be more than ICAs.

***Theorem 2.3*** **(Sparsity Operator is Contraction)** If we denote the sparse operator as $\mathcal{S}: \mathbb{R}^{S \times T} \to \mathbb{R}^{S \times T}$, we have $\|\mathcal{S}\| < \infty$.

***Proof***: according to the definition of operator norm (Rudin, 1973), $\|\mathcal{S}\| \leq \sup \frac{\|\mathcal{S}X\|}{\|X\|}$; obviously, $\|\mathcal{S}X\|$ and $\|X\|$ is bounded, since both of norms are based on finite dimensional matrix. And if we denote:

$$X = \begin{bmatrix} a_1 \\ a_2 \\ \vdots \\ a_{n-1} \\ a_n \end{bmatrix} \quad \mathcal{S}X = \begin{bmatrix} a_1 \\ 0 \\ \vdots \\ a_{n-1} \\ a_n \end{bmatrix} \tag{B.12}$$

Without loss of generality, and based on Lemma 1.2, we calculate the $\ell_2$ norm, and we have:
$$s = \|\mathcal{S}\| \leq \sup \frac{\|\mathcal{S}X\|}{\|X\|} = \frac{\sum_{i=1}^{k}(a_i)^2}{\sum_{i=1}^{n}(a_i)^2} \tag{B.13}$$

Since $k < n$,
$$s = \|\mathcal{S}\| < 1 \tag{B.14}$$

This inequality demonstrates that $\|\mathcal{S}\|$ is contraction operator.

***Theorem 2.4*** **(Random Initialization Operator is bounded)** If we denote the sparse operator as $\mathcal{J}: \mathbb{R}^{S \times T} \to \mathbb{R}^{S \times T}$, we have $\|\mathcal{M}\| < \infty$.

***Proof***: according to the definition of operator norm (Rudin 1973), $\|\mathcal{J}\| \leq \sup \frac{\|MX\|}{\|X\|}$; obviously, $\|\mathcal{J}X\|$ and $\|X\|$ is bounded, since both of norms are based on finite dimensional matrix. And if we denote:

$$X = \begin{bmatrix} a_1 \\ a_2 \\ \vdots \\ a_{n-1} \\ a_n \end{bmatrix} \quad \mathcal{J}X = \begin{bmatrix} b_1 \\ b_2 \\ \vdots \\ b_{n-1} \\ b_n \end{bmatrix} \quad \|X\| < \infty \quad \|\mathcal{J}X\| < \infty \tag{B.15}$$

Obviously, $\|\mathcal{J}\| < \infty$.

***Theorem 2.5*** **(Sparsity Operator is Contraction)** If we denote the sparse operator as $\mathcal{S}: \mathbb{R}^{S \times T} \to \mathbb{R}^{S \times T}$, we have $\|\mathcal{S}\| < \infty$.

***Proof***: according to the definition of operator norm (Rudin, 1973), $\|\mathcal{S}\| \leq sup \frac{\|\mathcal{S}X\|}{\|X\|}$; obviously, $\|\mathcal{S}X\|$ and $\|X\|$ is bounded, since both of norms are based on finite dimensional matrix. And if we denote:

$$X = \begin{bmatrix} a_1 \\ a_2 \\ \vdots \\ a_{n-1} \\ a_n \end{bmatrix} \quad \mathcal{S}X = \begin{bmatrix} a_1 \\ 0 \\ \vdots \\ a_{n-1} \\ a_n \end{bmatrix} \tag{B.16}$$

Without loss of generality, and based on Lemma 1.2, we calculate the $\ell_2$ norm, and we have:

$$s = \|\mathcal{S}\| \leq sup \frac{\|\mathcal{S}X\|}{\|X\|} = \frac{\sum_{i=1}^{k}(a_i)^2}{\sum_{i=1}^{n}(a_i)^2} \tag{B.17}$$

Since $k < n$,

$$s = \|\mathcal{S}\| < 1 \tag{B.18}$$

This inequality demonstrates that $\|\mathcal{S}\|$ is contraction operator.